\documentclass[10pt,twocolumn,letterpaper]{article}

\usepackage{iccv}              

\definecolor{iccvblue}{rgb}{0.21,0.49,0.74}
\usepackage[pagebackref,breaklinks,colorlinks,allcolors=iccvblue]{hyperref}
\usepackage{algorithm}
\usepackage{algpseudocode}

\def\paperID{6923} 
\def\confName{ICCV}
\def\confYear{2025}

\newcommand{\scheme}{Geminio}

\title{\scheme{}: Language-Guided Gradient Inversion Attacks in Federated Learning}

\author{
Junjie Shan
\quad Ziqi Zhao
\quad Jialin Lu
\quad Rui Zhang
\quad Siu Ming Yiu
\quad Ka-Ho Chow\thanks{Corresponding Author: kachow@cs.hku.hk}\\
School of Computing and Data Science\\
The University of Hong Kong, Hong Kong SAR, China\\
}

\begin{document}
\maketitle
\begin{abstract}
	Foundation models that bridge vision and language have made significant progress. While they have inspired many life-enriching applications, their potential for abuse in creating new threats remains largely unexplored. In this paper, we reveal that vision-language models (VLMs) can be weaponized to enhance gradient inversion attacks (GIAs) in federated learning (FL), where an FL server attempts to reconstruct private data samples from gradients shared by victim clients. Despite recent advances, existing GIAs struggle to reconstruct high-resolution images when the victim has a large local data batch. One promising direction is to focus reconstruction on valuable samples rather than the entire batch, but current methods lack the flexibility to target specific data of interest. To address this gap, we propose \scheme{}, the first approach to transform GIAs into semantically meaningful, targeted attacks. It enables a brand new privacy attack experience: attackers can describe, in natural language, the data they consider valuable, and Geminio will prioritize reconstruction to focus on those high-value samples. This is achieved by leveraging a pretrained VLM to guide the optimization of a malicious global model that, when shared with and optimized by a victim, retains only gradients of samples that match the attacker-specified query. \scheme{} can be launched at any FL round and has no impact on normal training (i.e., the FL server can steal clients' data while still producing a high-utility ML model as in benign scenarios).	Extensive experiments demonstrate its effectiveness in pinpointing and reconstructing targeted samples, with high success rates across complex datasets and large batch sizes with resilience against defenses.
\end{abstract}

\section{Introduction}\label{sec:intro}

\begin{figure}
	\centering
	\includegraphics[width=0.93\linewidth]{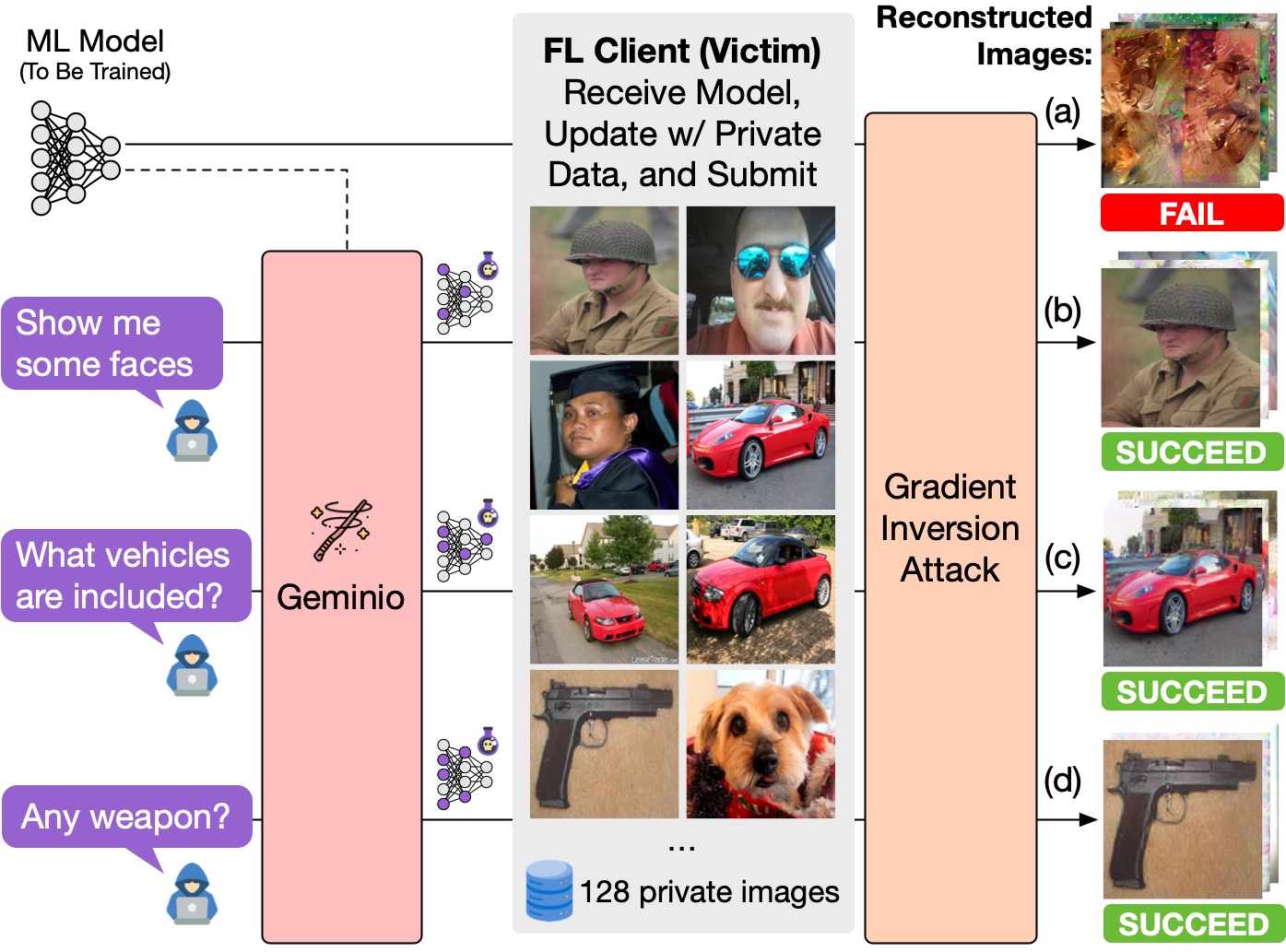}\vspace{-0.6em}
	\caption{\scheme{} enables the attacker (a malicious server in FL) to describe what kind of data is valuable to them and prioritize gradient inversion to recover those images from a large data batch.}\label{fig:showcase}\vspace{-1.15em}
\end{figure}

Federated learning (FL) is a privacy-enhancing technology for training machine learning models on data distributed across multiple clients~\cite{mcmahan2017communication,wen2023survey}. By enabling clients to share gradients rather than raw data with a coordinating server, FL has demonstrated transformative potential in privacy-sensitive domains~\cite{gartner,Liu_2021_CVPR,zhang2022federated}. However, FL is vulnerable to various malicious attacks, with gradient inversion attacks (GIAs) posing a particularly critical threat~\cite{zhang2023survey,huang2021evaluating}. These attacks enable a malicious FL server to reconstruct private data samples from the gradients a victim client shares.


GIAs face a longstanding challenge in reconstructing images from gradients produced by a large batch of data~\cite{fang2024privacy,tabassum2024efficiency}, an issue that has become increasingly relevant with improved hardware and training strategies~\cite{smith2018don} that enable larger batch sizes.
This limitation exists because GIAs rely on searching for data that reproduces the victim-submitted gradients, and the search space expands exponentially with batch size. While some approaches incorporate image priors~\cite{geiping2020inverting} to facilitate the search, a performance gap still remains, as illustrated in Figure~\ref{fig:showcase}a, with images reconstructed from a batch of 128 samples. Thus, much research has focused on narrowing the scope to reconstruct only a subset of samples. While promising, existing methods lack a semantically meaningful way for the adversary to specify which samples are preferred and can only target, e.g., outliers~\cite{wen2022fishing} or images with particular brightness levels~\cite{fowl2021robbing}. This raises an intriguing question: can reconstruction efforts be prioritized toward the data samples that truly matter to the adversary? If so, how can we allow the adversary to specify their preferences in a meaningful, flexible, and generic way?

In this paper, we empower GIAs with a natural language interface and propose Geminio. It enables the FL server to provide a natural language query describing the data of interest, allowing Geminio to prioritize and reconstruct matching data samples. Taking the batch of images from a victim's mobile phone in Figure~\ref{fig:showcase} as an example, the adversary could submit queries like (b) ``show me some faces" to retrieve images containing faces to see the victim or their friends, (c) ``what vehicles are included?" to identify cars associated with the victim, or (d) ``any weapon?" to detect if the victim owns a weapon. The query does not need to relate to the FL system's ML task. The FL server using Geminio can steal data at any FL round (even the first) while still producing a high-utility ML model, as in benign scenarios. By prioritizing reconstruction efforts, Geminio can pinpoint and retrieve targeted samples from large batches, offering high flexibility in defining valuable data. This capability is achieved by abusing pretrained vision-language models (VLMs)~\cite{radford2021learning} to help craft a malicious global model. When shared with and optimized by the victim, gradients become dominated by samples that match the query. Existing reconstruction optimization algorithms~\cite{zhu2019deep,geiping2020inverting,yin2021see,ye2024high} can consume such gradients to recover high-quality, targeted data.

Our main contributions are summarized as follows. First, we explore the abuse of pretrained VLMs to bridge the gap in gradient inversion, enabling semantically meaningful, targeted attacks. We investigate the first natural language interface for the adversary to describe the data samples that truly matter and prioritize them for reconstruction. Second, we propose Geminio, which exploits a VLM to reshape the loss surface of a malicious global model so that once optimized by the victim, the gradients are dominated by the samples matching the query. This method complements existing reconstruction optimizations and can augment them as targeted attacks. Third, we reveal the limitations of current defenses, discuss potential design improvements, and highlight their shortcomings to motivate future work. 
Experiments were conducted across three datasets, five attack methods, four defense mechanisms, and various configurations to assess the threat posed by Geminio. 
The source code of Geminio is available at \url{https://github.com/HKU-TASR/Geminio}.

\section{Background}\label{sec:related-work}
\textbf{Federated Learning.} Let $F_{\boldsymbol{\theta}}$ be the ML model trained via FL with a loss function $\mathcal{L}$. At each learning round $t$, the FL server sends the current global model parameters $\boldsymbol{\theta}_t$ to selected FL clients. Under the FedSGD protocol~\cite{mcmahan2017communication}, each client $i$ samples a data batch $\boldsymbol{\mathcal{B}}^i_t$, having pairs of input $\boldsymbol{x}$ and label $y$, from its private dataset to optimize the received model and submits the gradients
\begin{equation}\label{eq:fedsgd}
	\boldsymbol{\mathcal{G}}(\boldsymbol{\mathcal{B}}^i_t;\boldsymbol{\theta}_t)=\frac{1}{\vert\boldsymbol{\mathcal{B}}^i_t\vert}\sum_{(\boldsymbol{x}, y)\in\boldsymbol{\mathcal{B}}^i_t}\nabla_{\boldsymbol{\theta}_t}\mathcal{L}(F_{\boldsymbol{\theta}_t}(\boldsymbol{x}); y)
\end{equation}
to the server. Then, the server aggregates the gradients submitted by the clients to update the global model parameters for the next round. The FL protocol has different variations, such as selecting all or a random subset of clients to participate in each round or running several data batches locally before submitting the gradients to the server~\cite{mcmahan2017communication,luo2025harmless}.

\noindent\textbf{Gradient Inversion Attacks.} The FL server that receives the gradients $\boldsymbol{\mathcal{G}}(\boldsymbol{\mathcal{B}}^i_t;\boldsymbol{\theta}_t)$ from a participating client $i$ at round $t$ can reconstruct the private data batch $\boldsymbol{\mathcal{B}}^i_t$ via gradient inversion attacks. As the batch size should be transparent to the FL server for proper aggregation, it can randomly initialize a batch of data samples $\bar{\boldsymbol{\mathcal{B}}}$ and use the global model parameters $\boldsymbol{\theta}_t$ shared with the victim at the beginning of the learning round for reconstruction optimization:
\begin{equation}\label{eq:re-opt}
	\bar{\boldsymbol{\mathcal{B}}}^*=\underset{\bar{\boldsymbol{\mathcal{B}}}}{\mathrm{argmin}}\;\big[{\delta}(\boldsymbol{\mathcal{G}}(\boldsymbol{\mathcal{B}}^i_t;\boldsymbol{\theta}_t),\boldsymbol{\mathcal{G}}(\bar{\boldsymbol{\mathcal{B}}};\boldsymbol{\theta}_t)) + \mathcal{R}(\bar{\boldsymbol{\mathcal{B}}})\big],
\end{equation}
where $\delta$ is a distance function that measures the dissimilarity between two sets of gradients, and $\mathcal{R}$ is a regularization function. The overarching idea is to optimize those random data samples in such a way that they can reproduce the gradients shared by the victim client. 

\begin{figure*}
	\centering
	\includegraphics[width=\linewidth]{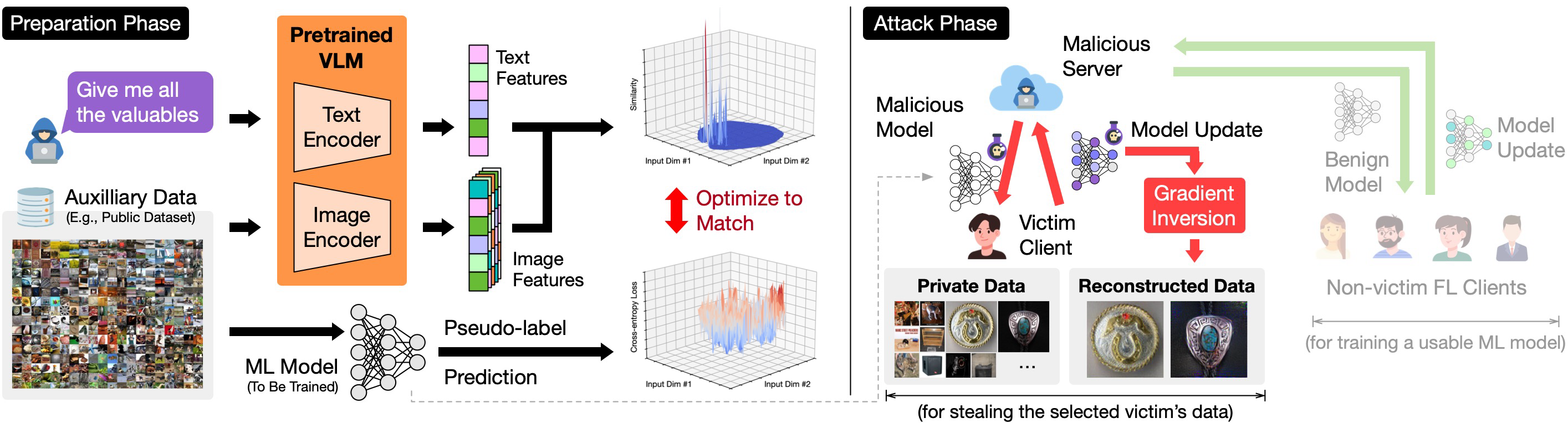} \vspace{-1em}
	\caption{\scheme{} begins with a preparation phase. It receives a query from the attacker and uses a pretrained VLM and an auxiliary, unlabeled dataset to reshape the loss surface of the malicious global model. During the attack phase, at an FL round, the FL server (i) sends the malicious model to the victim client and (ii) maintains a group of non-victim clients to train a usable ML model as usual. The gradients (model updates) from the victim can be fed to existing reconstruction optimization methods to recover images matching the query. }\label{fig:overview}\vspace{-1em}
\end{figure*}

\subsection{Related Work}


\noindent\textbf{Reconstruction Optimization.} The reconstruction optimization in Equation~\ref{eq:re-opt} has been the primary focus of GIA advancements. These improvements target (i) enhanced distance functions, such as Euclidean distance~\cite{zhu2019deep} and cosine similarity~\cite{geiping2020inverting,ye2024high} to measure gradient alignment, and (ii) refined regularization methods to incorporate prior knowledge like spatial smoothness~\cite{geiping2020inverting,yin2021see}. Such approaches have successfully extended GIAs to support high-resolution image reconstructions, previously limited to toy datasets like MNIST~\cite{zhu2019deep}. However, these methods attempt to recover the entire private batch, struggling with practical batch sizes due to the vast search space involved~\cite{shi2024dealing}. 
Even if uncommon modifications to the neural architecture are allowed, such as replacing all activation functions with a sigmoid function, as in CI-Net~\cite{zhang2023generative}, it can only support a batch size of $32$ on ImageNet.

\noindent\textbf{Narrowing the Reconstruction Scope.} Recognizing this limitation, recent work has explored reconstructing only a subset of data samples in the private batch by manipulating the global model shared with the victim client. Abandon~\cite{boenisch2023curious}, Robbing~\cite{fowl2021robbing}, GradViT~\cite{Hatamizadeh_2022_CVPR}, LOKI~\cite{zhao2024loki}, and SEER~\cite{garov2024hiding} introduce special neural architectures to retain only the gradients of selected private samples during the victim’s local training. These “trapped” samples are either random or satisfy simple conditions, such as brightness levels or average color intensity. Although this narrows the reconstruction scope, it gives the adversary almost no meaningful control over which specific samples are recovered. Also, the unusual neural architecture can be suspicious, even if it enables analytical reconstructions. Fishing~\cite{wen2022fishing} and GradFilt~\cite{zhang2024gradfilt} improve this by setting certain model parameters at the output layer to very large values, causing the gradients to be dominated by one or all samples of a particular class, after which reconstruction optimization algorithms can be employed to recover them. However, this control remains restrictive, as the adversary can neither specify finer-grained sample characteristics within a class nor define conditions irrelevant to the FL system’s ML task, not to mention the unnaturally large parameter values can be easily detected. Imperio~\cite{chow2024imperio} is the first to propose natural language-guided backdoor attacks. However, there has been no similar approach for GIA. To address these limitations, this paper introduces Geminio, offering the first natural language interface for targeted GIAs.

\noindent\textbf{Defenses.} Encryption-based methods, such as homomorphic encryption~\cite{fang2021privacy}, have been proposed to secure gradient confidentiality but are often computationally prohibitive~\cite{zhang2020batchcrypt} or can be circumvented if an active FL server modifies the FL protocol~\cite{boenisch2023reconstructing}. Gradient obfuscation techniques, such as differential privacy~\cite{wei2020federated} and gradient pruning~\cite{zhang2023preserving}, allow FL clients to protect their data proactively~\cite{10.1007/978-3-030-58951-6_27,wei2023securing,wei2023model}. However, as our experiments reveal, these defenses fail to mitigate the privacy threats posed by Geminio.

\subsection{Threat Model}
Consistent with prior studies~\cite{wen2022fishing,zhang2024gradfilt}, we consider an FL server acting as an active adversary who (i) can modify the model parameters before sharing them with FL clients but not altering the neural architecture, (ii) can read the gradients submitted by a victim client and attempt to reconstruct private data samples from them, (iii) can provide a natural language description of the characteristics of data it deems valuable, and (iv) possesses an auxiliary, unlabeled image dataset that may originate from a completely different domain (e.g., public datasets like ImageNet~\cite{imagenet} or images scraped from the Internet). The FL clients adhere to the FedSGD protocol, optimizing the received model with a batch of private data. We will consider other FL scenarios in Section~\ref{sec:experiments}, such as Geminio under FedAvg~\cite{mcmahan2017communication} and client-side defenses. As to be discussed, even though this threat model requires sharing malicious global model parameters, it is realistic and non-trivial to be detected by victims because a typical FL server controls the client selection process, and Geminio-generated model parameters are stealthy. 

\section{Methodology}\label{sec:methodology}
Figure~\ref{fig:overview} gives an overview of Geminio, which consists of two phases.
During the preparation phase, which may occur even before FL begins, Geminio takes a query $\mathcal{Q}$ (e.g., ``show me all valuables") from the adversary, randomly initializes a malicious global model, and optimizes its parameters $\boldsymbol{\Theta}_{\mathcal{Q}}$ based on the query. 
During the attack phase, at an FL round, while the FL server follows the standard FL protocol to communicate with a group of non-victim clients for training a usable ML model, it sends the malicious global model parameters to the victim. The victim then optimizes these parameters with its private data batch $\boldsymbol{\mathcal{B}}$ and uploads the resulting gradients $\boldsymbol{\mathcal{G}}(\boldsymbol{\mathcal{B}};\boldsymbol{\Theta}_{\mathcal{Q}})$ (from Equation~\ref{eq:fedsgd}) to the FL server. Afterward, any existing reconstruction optimization method can be directly applied to recover private samples relevant to the query (e.g., the jewelry retrieved by HFGradInv~\cite{ye2024high}). Note that the FL server only uses the gradients submitted by non-victim clients for aggregation. As a result, it can produce a usable ML model with the same utility as benign scenarios.

The overarching idea of Geminio is to craft a malicious global model such that those private samples in the victim's data batch matching the query will dominate the submitted gradients. Consider the scenario that only one private sample $(\boldsymbol{x}_\text{target}, y_\text{target})\in\boldsymbol{\mathcal{B}}$ matches the query as an example; the malicious global model should behave as follows when being optimized by the victim client: 
$\vert\vert\nabla_{\boldsymbol{\Theta}_\mathcal{Q}}\mathcal{L}(F_{\boldsymbol{\Theta}_\mathcal{Q}}(\boldsymbol{x}); y)\vert\vert\ll\vert\vert\nabla_{\boldsymbol{\Theta}_\mathcal{Q}}\mathcal{L}(F_{\boldsymbol{\Theta}_\mathcal{Q}}(\boldsymbol{x}_{\text{target}}); y_{\text{target}})\vert\vert$ for all $\boldsymbol{x}\neq\boldsymbol{x}_{\text{target}}$.
Then, the victim-submitted gradients become $\boldsymbol{\mathcal{G}}(\boldsymbol{\mathcal{B}};\boldsymbol{\Theta}_{\mathcal{Q}})\approx\frac{1}{\vert\boldsymbol{\mathcal{B}}\vert}\nabla_{\boldsymbol{\Theta}_\mathcal{Q}}\mathcal{L}(F_{\boldsymbol{\Theta}_\mathcal{Q}}(\boldsymbol{x}_{\text{target}}); y_{\text{target}})$, and any existing reconstruction optimization method (see Equation~\ref{eq:re-opt}) will recover $\boldsymbol{x}_{\text{target}}$ from them. To achieve such a behavior, instead of directly optimizing how the global model should produce gradients, which involves second-order derivatives and is highly unstable, we could exploit the property that the per-sample gradient magnitude $\vert\vert\nabla_{\boldsymbol{\Theta}_\mathcal{Q}}\mathcal{L}(F_{\boldsymbol{\Theta}_\mathcal{Q}}(\boldsymbol{x}); y)\vert\vert$ is proportional to the per-sample loss value $\mathcal{L}(F_{\boldsymbol{\Theta}_\mathcal{Q}}(\boldsymbol{x}); y)$. Geminio's objective is to craft a malicious model that amplifies the loss value of matched samples while suppressing the others. 

Crafting such a malicious model for a given query requires two key components. First, we need a supervisor to guide how the model should react to an input image. In Section~\ref{sec:reshape}, we introduce a VLM-guided approach that leverages the text-image association capabilities of VLMs as a learning signal. Second, we need an auxiliary training dataset. In Section~\ref{sec:label}, we present a VLM-guided approach that enables Geminio to utilize unlabeled data that is readily available (e.g., public datasets or web-scraped images).

%
%

\subsection{VLM-Guided Loss Surface Reshaping}\label{sec:reshape}
Given an auxiliary dataset $\boldsymbol{\mathcal{A}}$, Geminio exploits a pretrained VLM to measure the similarity between each auxiliary image and the query. The top 3D surface plot in Figure~\ref{fig:overview} shows the similarity surface as a function of auxiliary images (projected onto a 2D space by PCA). Some images align with the query well (red), while others have close-to-zero relatedness (blue). Since a randomly initialized, untrained global model has a large loss value across different images (see the bottom 3D surface plot in Figure~\ref{fig:overview}), we need to train this model to have a loss surface matching the aforementioned similarity surface. Then, samples irrelevant to the query will have a negligible loss, while the relevant ones will dominate. 

\begin{figure}
	\centering
	\begin{subfigure}[b]{0.49\linewidth}
		\centering
		\includegraphics[width=0.8\linewidth]{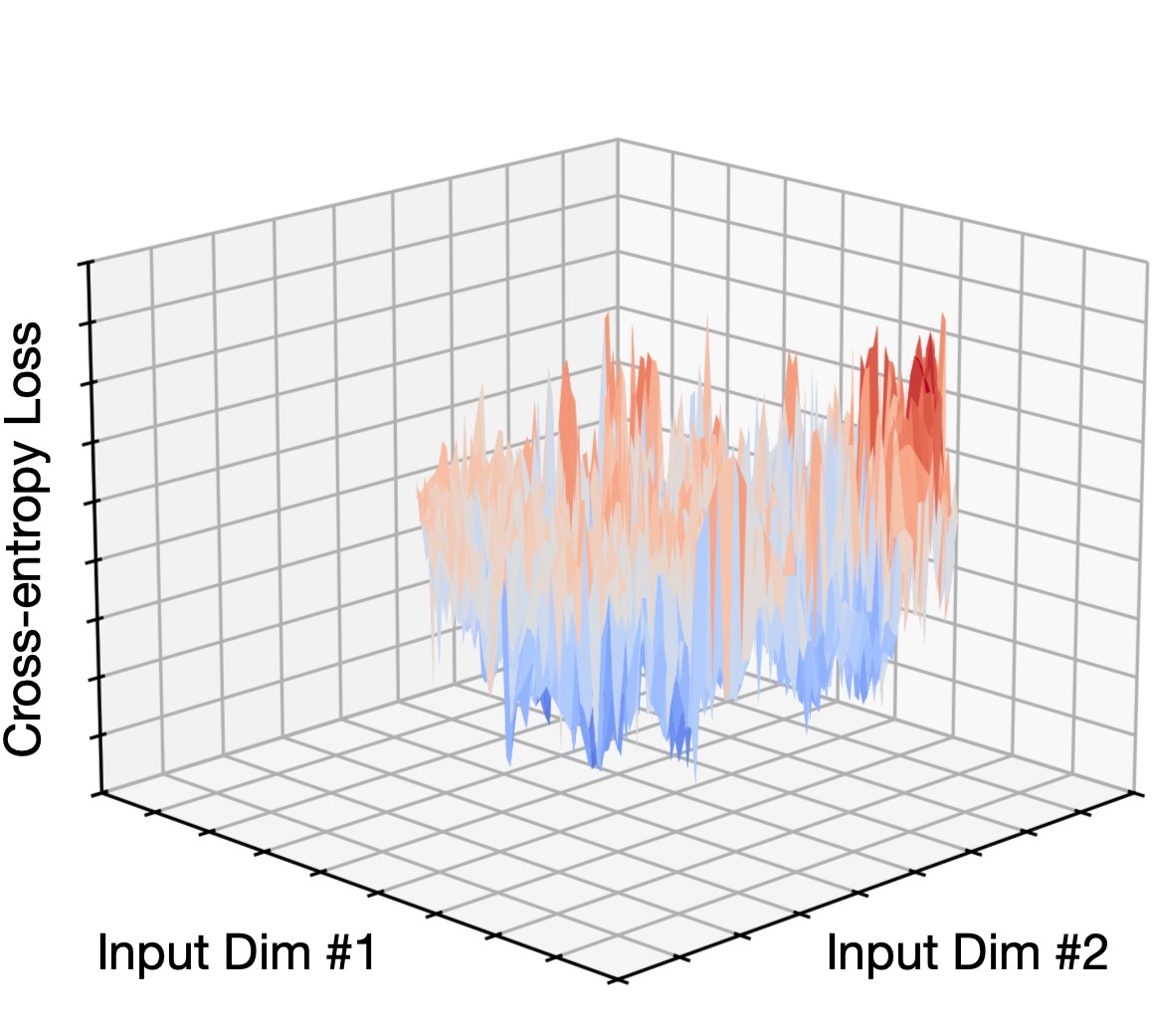} 
		\caption{Original Global Model}\label{fig:loss-landscape-a}\vspace{-0.5em}
	\end{subfigure}
	\begin{subfigure}[b]{0.49\linewidth}
		\centering
		\includegraphics[width=0.8\linewidth]{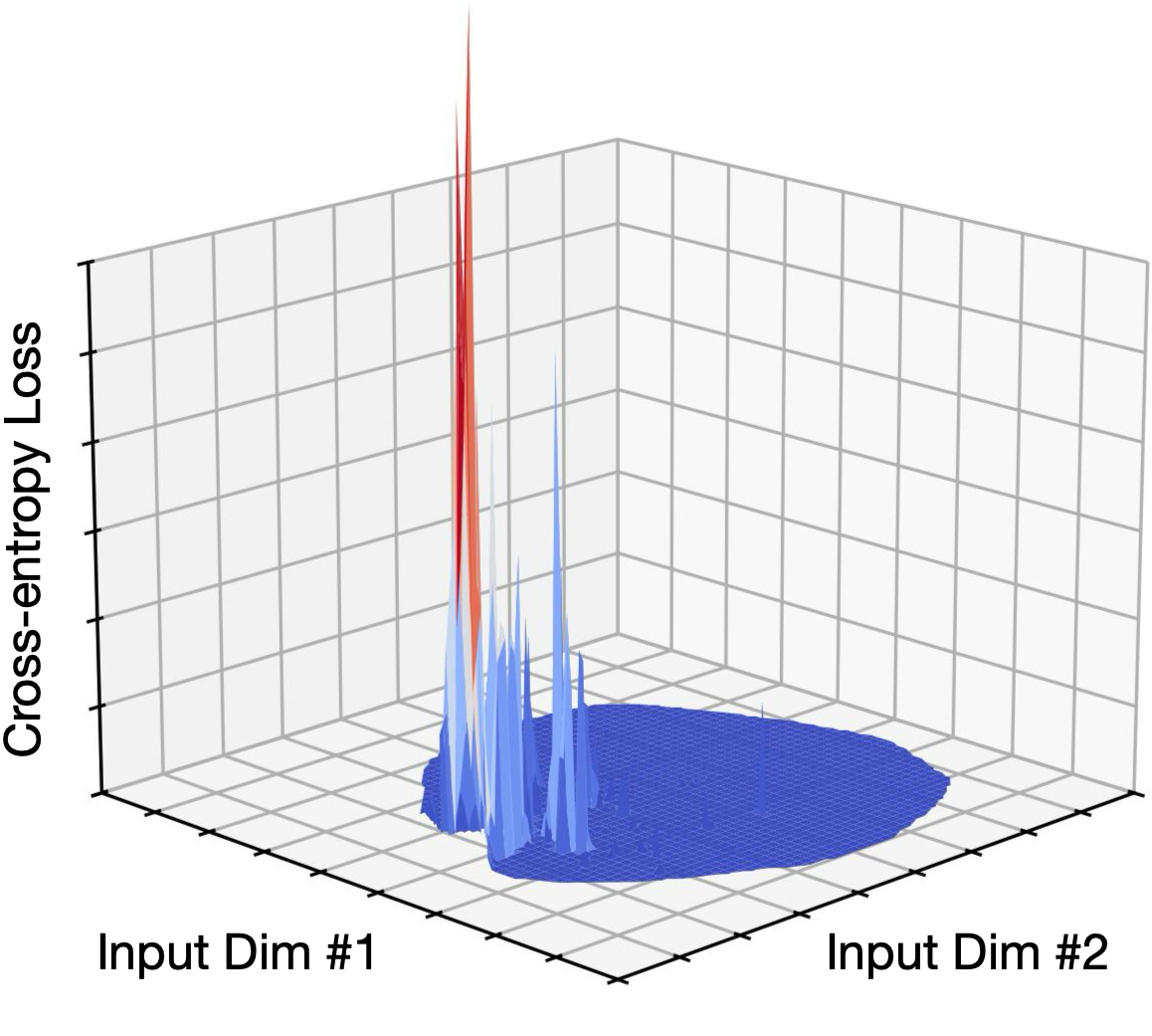} 
		\caption{Reshaped by \scheme{}}\label{fig:loss-landscape-b}\vspace{-0.5em}
	\end{subfigure}
	\caption{\scheme{} reshapes the loss landscape of the model such that only the samples matching the query will have an amplified loss to dominate gradients for targeted reconstruction.}\label{fig:loss-landscape}\vspace{-1em}
\end{figure}

A VLM comprises two components~\cite{radford2021learning}: an image encoder $\mathcal{V}_\text{image}$ and a text encoder $\mathcal{V}_\text{text}$. They can project image and text data onto a latent space that those similar will collocate. For an auxiliary sample $(\boldsymbol{x}, y)\in\boldsymbol{\mathcal{A}}$, we can calculate its similarity with the query $\mathcal{Q}$: $s(\boldsymbol{x};\mathcal{Q})=\mathcal{V}_{\text{image}}(\boldsymbol{x})^{\intercal}\mathcal{V}_{\text{text}}(\mathcal{Q})$.
Based on the similarity score, we propose to train the malicious global model parameters $\boldsymbol{\Theta}_{\mathcal{Q}}$ with the following routine. At each iteration, we sample a batch of auxiliary data $\boldsymbol{\mathcal{B}}_{\text{aux}}\subset\boldsymbol{\mathcal{A}}$ and calculate the probability of each auxiliary image $(\boldsymbol{x}, y)\in\boldsymbol{\mathcal{B}}_{\text{aux}}$ being aligned with the query, normalized across the batch via a softmax function:
\begin{equation}
	\alpha(\boldsymbol{x};\mathcal{Q}, \boldsymbol{\mathcal{B}}_{\text{aux}})=\frac{\exp(s(\boldsymbol{x};\mathcal{Q}))}{\sum_{(\boldsymbol{x}', y')\in\boldsymbol{\mathcal{B}}_{\text{aux}}}\exp(s(\boldsymbol{x}';\mathcal{Q}))}.
\end{equation}
The batch-wise normalization offers information about how one sample is more aligned with the query than another sample. Then, we can train the malicious global model parameters by minimizing
\begin{equation}\label{eq:geminio}\small
	\begin{split}
		&\mathcal{L}_{\text{Geminio}}(\boldsymbol{\mathcal{B}}_{\text{aux}}; F_{\boldsymbol{\Theta}_{\mathcal{Q}}}, \mathcal{Q})\\
		=&\frac{\sum_{(\boldsymbol{x}, y)\in\boldsymbol{\mathcal{B}}_{\text{aux}}}\mathcal{L}(F_{\boldsymbol{\Theta}_{\mathcal{Q}}}(\boldsymbol{x}); y)(1-\alpha(\boldsymbol{x};\mathcal{Q}, \boldsymbol{\mathcal{B}}_{\text{aux}}))}{\vert\boldsymbol{\mathcal{B}}_{\text{aux}}\vert\sum\limits_{(\boldsymbol{x}', y')\in\boldsymbol{\mathcal{B}}_{\text{aux}}}\mathcal{L}(F_{\boldsymbol{\Theta}_{\mathcal{Q}}}(\boldsymbol{x}'); y')(1-\alpha(\boldsymbol{x}';\mathcal{Q}, \boldsymbol{\mathcal{B}}_{\text{aux}}))}.
	\end{split}
\end{equation}
Intuitively, each per-sample loss is associated with a scaling factor (the coefficient). For an auxiliary sample that has a strong alignment with the query, the corresponding term will be negligible since the coefficient $(1-\alpha(\boldsymbol{x}; \mathcal{Q}, \boldsymbol{\mathcal{B}}_{\text{aux}}))$ is small. In contrast, the term corresponding to an irrelevant auxiliary sample will have its magnitude amplified because of the large coefficient. In order for the malicious model to minimize Equation~\ref{eq:geminio}, it must learn to reduce the per-sample loss value of such irrelevant samples. The batch-wise normalization will then increase the per-sample loss value of matched samples. Geminio's training routine reshapes the loss surface (Figure~\ref{fig:loss-landscape-a}) to have an active response only for those matched samples (Figure~\ref{fig:loss-landscape-b}).

\begin{figure*}[h]
	\centering
	\begin{subfigure}[b]{0.33\linewidth}
		\centering
		\includegraphics[width=0.98\linewidth]{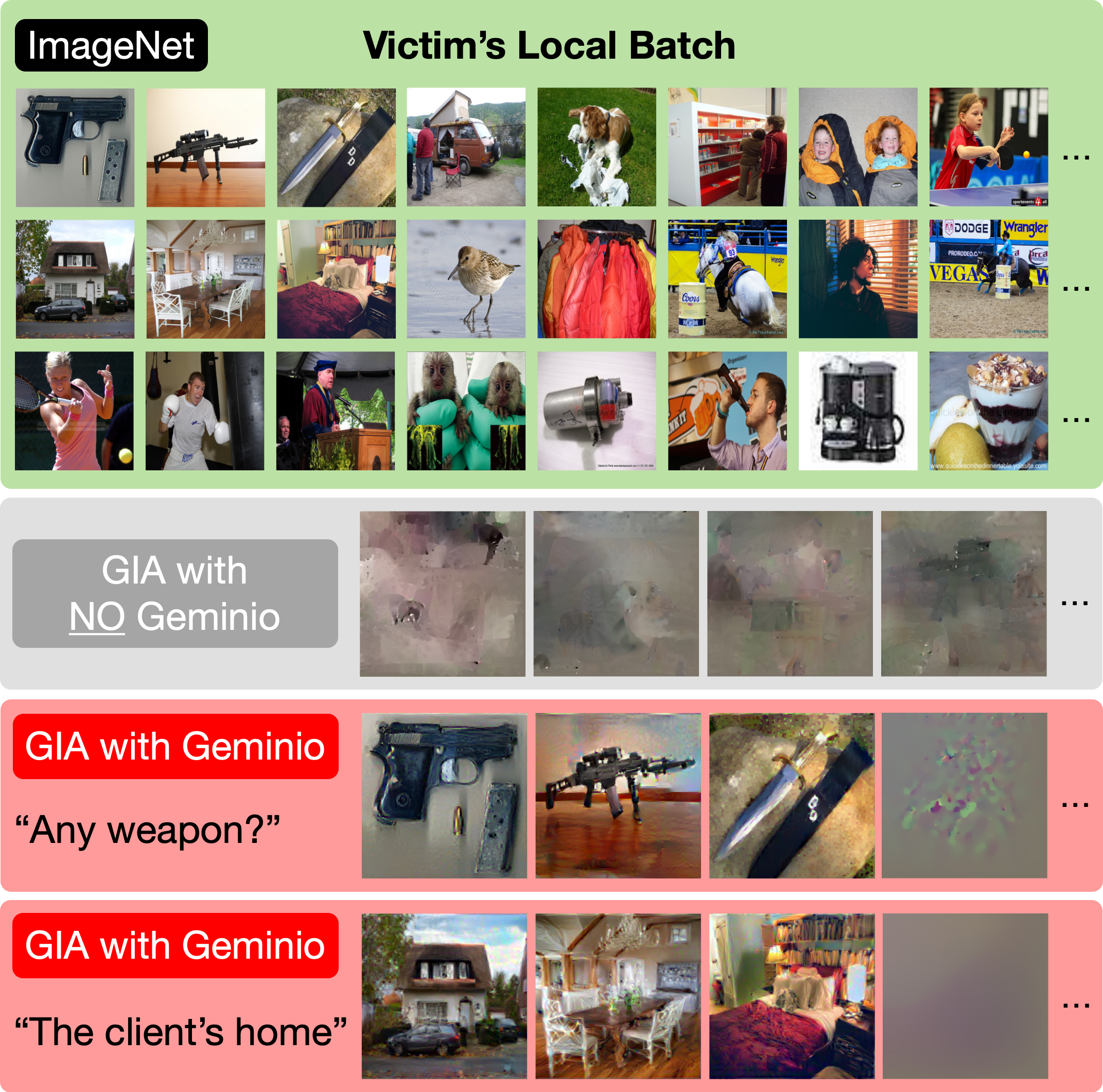}
		\caption{ImageNet}\label{fig:visual-imagenet}
	\end{subfigure}\hfill
	\begin{subfigure}[b]{0.33\linewidth}
		\centering
		\includegraphics[width=0.98\linewidth]{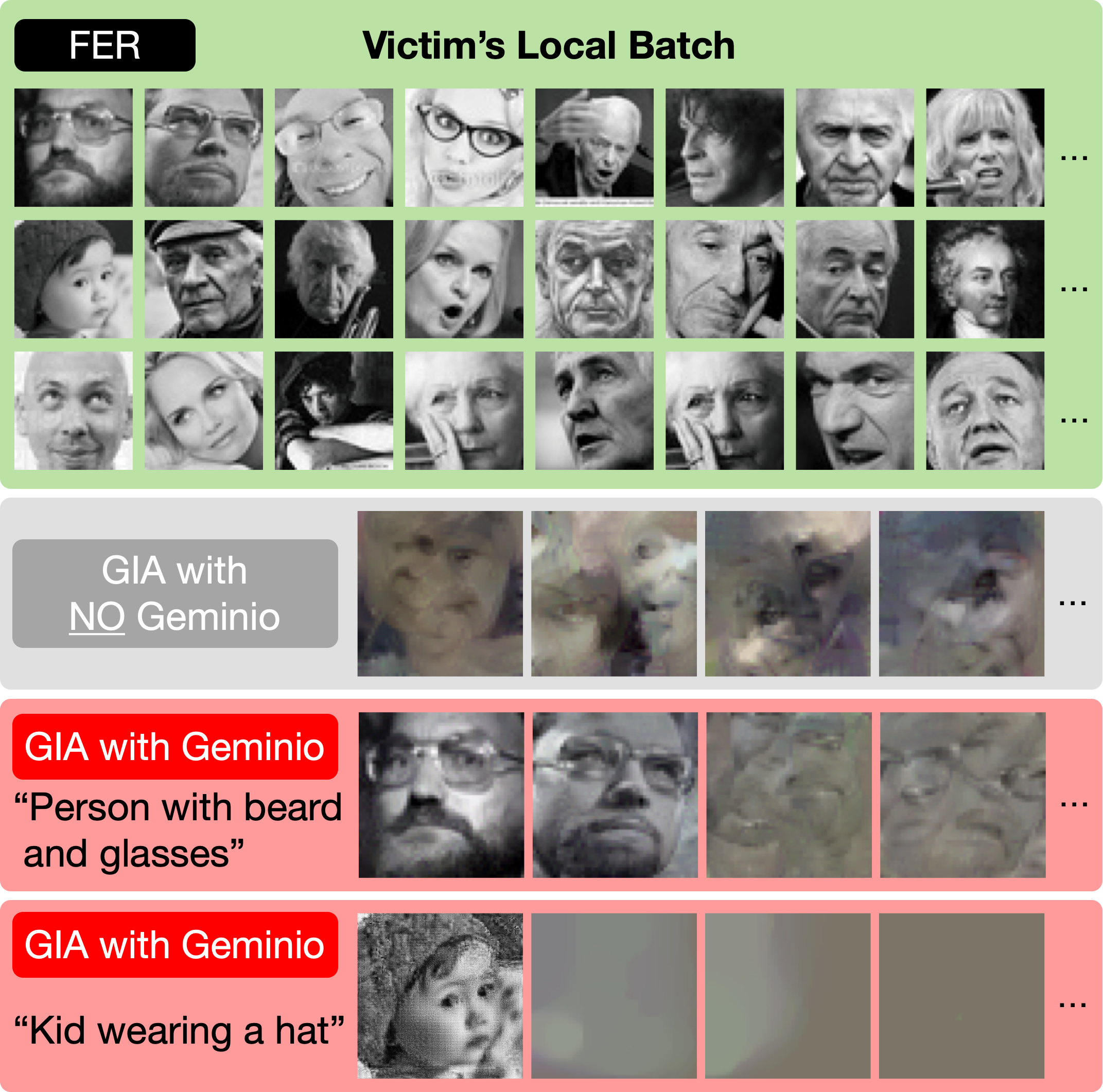}
		\caption{FER}\label{fig:visua-fer}
	\end{subfigure}\hfill
	\begin{subfigure}[b]{0.33\linewidth}
		\centering
		\includegraphics[width=0.98\linewidth]{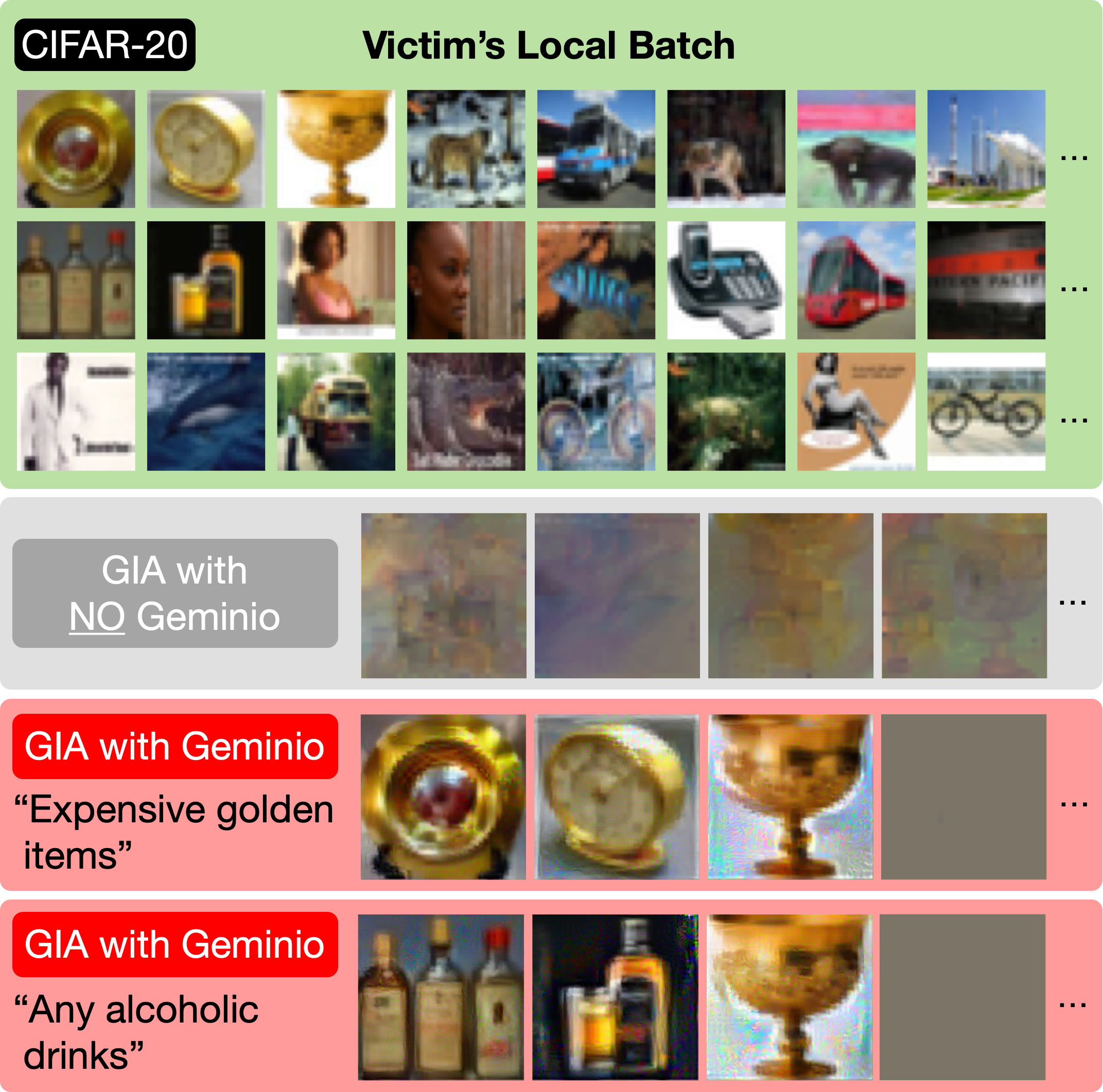}
		\caption{CIFAR-20}\label{fig:visua-cifar20}
	\end{subfigure}\vspace{-0.8em}
	\caption{\scheme{} can take task-agnostic queries from the attacker to achieve instance-level targeted reconstruction. While vanilla GIAs cannot recover recognizable images from a large batch, \scheme{} narrows down the reconstruction scope and successfully rebuilds high-fidelity images that match the attacker's queries (e.g., the handgun, rifle, and knife images given the query ``Any weapon?" in (a)).}\label{fig:visual}\vspace{-1em}
\end{figure*}

\subsection{VLM-Guided Auxiliary Label Generation}\label{sec:label}
\scheme{} requires an auxiliary dataset to optimize the malicious global model with Equation~\ref{eq:geminio}. Recall that the calculation of the per-sample loss value requires the ground-truth label of that input. We propose to abuse the pretrained VLM again to eliminate this requirement and enable Geminio with unlabeled samples, which can easily be collected. In particular, let the class names of the $K$-class classification task of FL be $[c_1, c_2, ..., c_K]$. We can generate a soft label for each auxiliary sample $\boldsymbol{x}$ by measuring the similarity of its image features and the text features of each class name. Formally, the soft label $\boldsymbol{y}=[y_1, y_2, ..., y_K]$ is a probability distribution, where $y_i$ represents the probability of $\boldsymbol{x}$ being classified as class $c_i$ and can be calculated by
\begin{equation}
	y_i = \frac{\mathcal{V}_{\text{image}}(\boldsymbol{x})^{\intercal}\mathcal{V}_{\text{text}}(c_i)}{\sum_{j=1}^{K}\mathcal{V}_{\text{image}}(\boldsymbol{x})^{\intercal}\mathcal{V}_{\text{text}}(c_j)}
\end{equation}
Using soft labels in the cross-entropy loss function, one could launch Geminio by simply using public datasets (e.g., ImageNet) or scraping images from the internet. The pseudocode of \scheme{} is provided in Appendix~\ref{supp:pseudo}.

\section{Empirical Evaluation}\label{sec:experiments}
We conduct extensive experiments to analyze \scheme{}'s broad applicability to different datasets (ImageNet~\cite{imagenet}, CIFAR-20~\cite{cifar100}, and FER~\cite{fer}), different neural architectures (ResNet~\cite{he2016deep}, MobileNet~\cite{Howard2019SearchingFM}, EfficientNet~\cite{Tan2021EfficientNetV2SM}, and ViT~\cite{dosovitskiy2021an}), and different FL scenarios (FedSGD and FedAvg). CIFAR-20 is equivalent to CIFAR-100 but uses 20 superclasses as labels. It provides ground truths to  evaluate \scheme{}'s task-agnostic targeted retrieval quantitatively.

By default, we consider an FL system that trains a ResNet34 model using FedSGD as the protocol with a batch size of 64. For \scheme{}, we use the pretrained CLIP~\cite{radford2021learning} to guide the optimization. The gradients are consumed by HFGradInv~\cite{ye2024high} to reconstruct private samples. Detailed setup is provided in Appendix~\ref{supp:setup}, and the source code is included as part of the supplementary materials to facilitate further research and reproducibility.

\noindent\textbf{Outline.} With additional analysis provided in the appendix, we would like to deliver three messages via the empirical studies in this section:
\begin{itemize}[leftmargin=*, noitemsep, topsep=0pt]
	\item The attacker can freely describe the data valuable to them and ``query" the victim's dataset for targeted reconstruction from a large batch of data. (Section~\ref{sec:exp-1})
	\item \scheme{} serves as a plugin to existing reconstruction optimization methods and is broadly applicable, even with limited access to auxiliary data. (Section~\ref{sec:exp-2})
	\item \scheme{} has a high survivability under various FL and defense scenarios. (Section~\ref{sec:exp-3})
\end{itemize}
\begin{figure*}
	\centering
	\begin{minipage}{.63\textwidth}
		\centering
		\includegraphics[width=\linewidth]{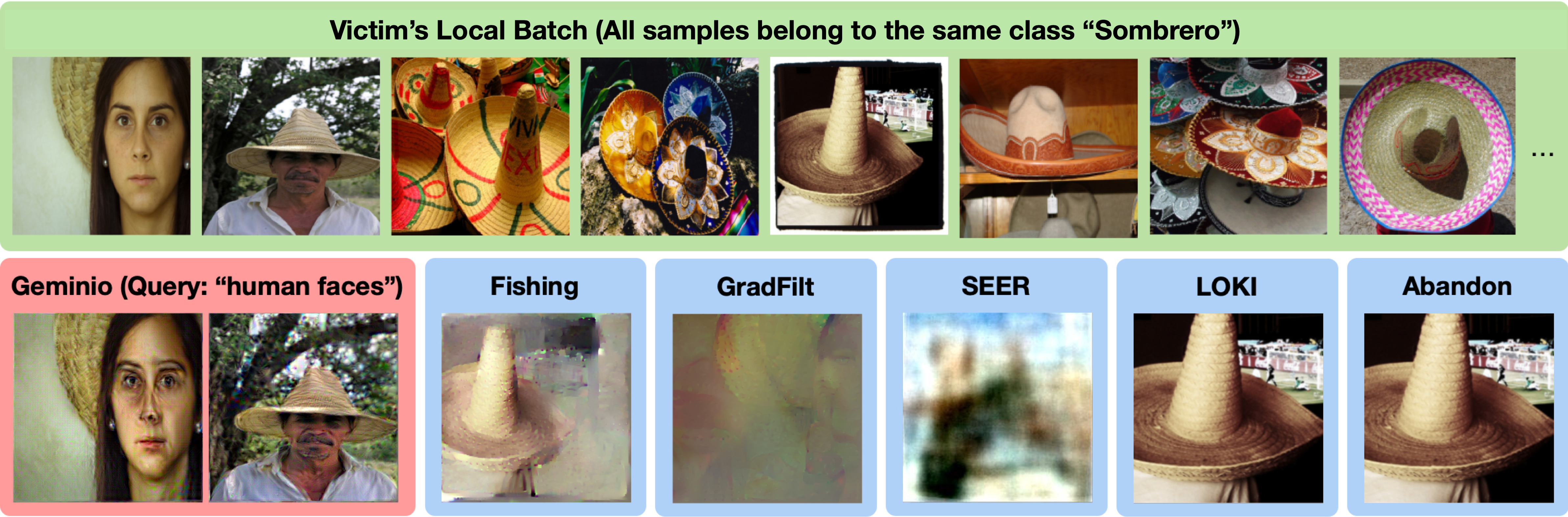}\vspace{-0.2em}
		\caption{\scheme{} is the only method that achieves task-agnostic, instance-level targeted reconstruction. Other approaches can only be class-level (Fishing and GradFilt) or can only consider semantic-irrelevant conditions (SEER, LOKI, and Abandon).}\label{fig:advance}\vspace{-0.8em}
	\end{minipage}\hfill
	\begin{minipage}{0.36\textwidth}
		\centering
		\begin{subfigure}[b]{0.49\linewidth}
			\centering
			\includegraphics[width=\linewidth]{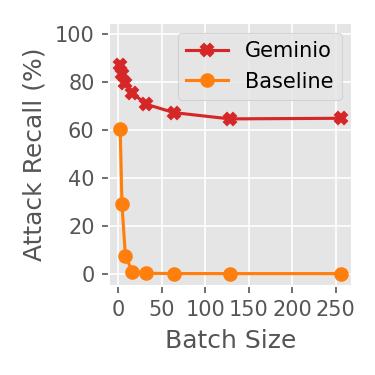}\vspace{-0.2em}
			\caption{Attack Recall}\label{fig:attack-recall}
		\end{subfigure}
		\begin{subfigure}[b]{0.49\linewidth}
			\centering
			\includegraphics[width=\linewidth]{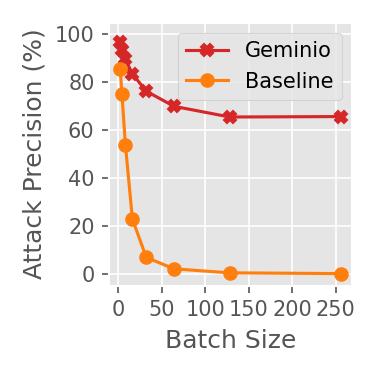}\vspace{-0.2em}
			\caption{Attack Precision}\label{fig:attack-precision}
		\end{subfigure}\vspace{-0.4em}
		\caption{\scheme{} remains effective even when the batch size used by the victim is large (e.g., 256). In comparison, the baseline method is virtually useless when the batch size is larger than 8. }\label{fig:batch-size}\vspace{-0.8em}
	\end{minipage}
\end{figure*}

\subsection{Task-agnostic, Targeted Reconstruction}\label{sec:exp-1}
\textbf{Qualitative Analysis.} To showcase \scheme{}'s targeted retrieval, Figure~\ref{fig:visual} provides three example batches of the victim's private data (top) from different datasets and the corresponding reconstructed images (bottom) for three cases: reconstruction with the vanilla GIA (1st row) and reconstruction with \scheme{} given two different queries (2nd and 3rd rows). 
First, while the vanilla GIA cannot produce recognizable images due to its failure to handle a large batch, \scheme{} narrows the reconstruction scope to the data samples that matter most and successfully recovers them with high fidelity.
Second, the recovered images match the attacker-provided queries. For instance, a curious attacker may submit a query ``Any weapon?" to understand whether the client is, e.g., a weapon enthusiast.  Among the $64$ images on ImageNet (Figure~\ref{fig:visual-imagenet}), only the first three contain weapons and are all successfully reconstructed. 
Similarly, considering the query ``Person with beard and glasses," while the first four images on FER (Figure~\ref{fig:visua-fer}) contain a person wearing glasses, only the first two are reconstructed, as the rest do not have a beard.
Third, queries can be irrelevant to the ML task. FER classifies facial images into one of the seven emotion expressions (e.g., happy, sad). Even though our example queries describe the appearance of individuals, the targeted reconstructions are successful. More visual examples are provided in Appendix~\ref{supp:visual}.


\noindent\textbf{Comparison with Existing Methods.} \scheme{}'s targeted reconstruction is unique and not achievable by existing methods. Figure~\ref{fig:advance} shows another batch of private data with images of the class ``Sombrero." Imagine that an attacker wants to recover images that contain human faces as a privacy-intrusive example. As shown in the first column (2nd row), \scheme{} successfully recovers the first two images in the victim's private data (1st row). The reconstructed images clearly reveal the facial features of the people with whom the client may interact. In contrast, other methods attempting to narrow the reconstruction scope cannot achieve the same goal. Fishing~\cite{wen2022fishing} can only return one random sample of a given class; GradFilt~\cite{zhang2024gradfilt} returns all samples of a given class; SEER~\cite{garov2024hiding}, LOKI~\cite{zhao2024loki}, and Abandon~\cite{boenisch2023curious} can only be random or specify semantic-irrelevant conditions (e.g., the brightness level). It is worth emphasizing that \scheme{} is instance-level. The matched data samples can belong to different classes. It is the only solution that achieves such a fine granularity.

\begin{figure*}
	\centering
	\begin{minipage}{.34\textwidth}
		\centering
		\includegraphics[width=\linewidth]{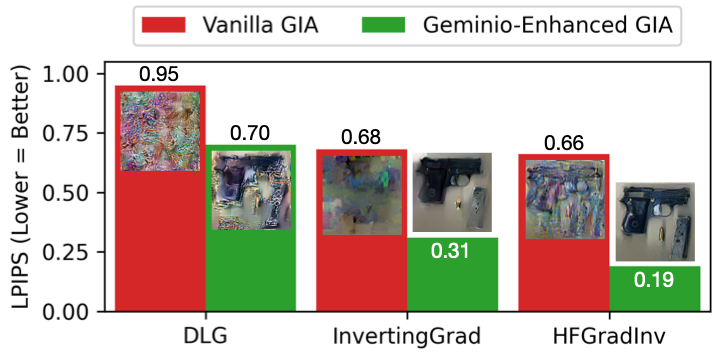}\vspace{-0.6em}
		\caption{\scheme{} complements existing reconstruction optimization methods, turns them into targeted attacks, and improves their reconstruction quality.}\label{fig:modular}
	\end{minipage}\hfill
	\begin{minipage}{.37\textwidth}
		\centering
		\includegraphics[width=\linewidth]{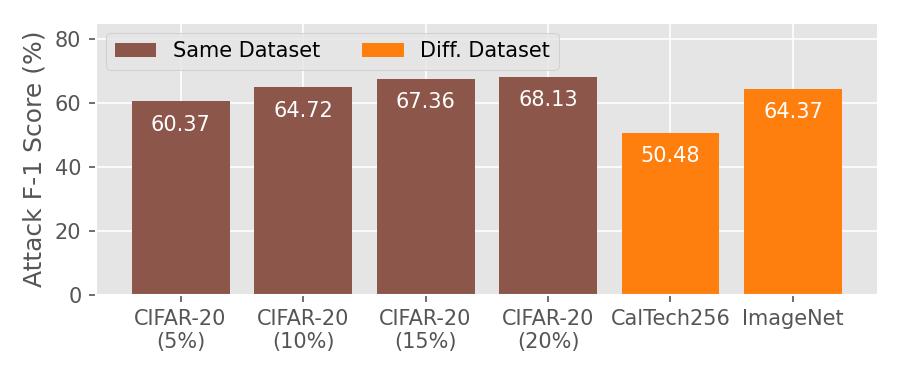}\vspace{-0.6em}
		\caption{A small number of samples from the same dataset or a different dataset can already drive \scheme{}.}\label{fig:aux-dataset}
	\end{minipage}\hfill
	\begin{minipage}{0.26\textwidth}
		\centering
		\includegraphics[width=\linewidth]{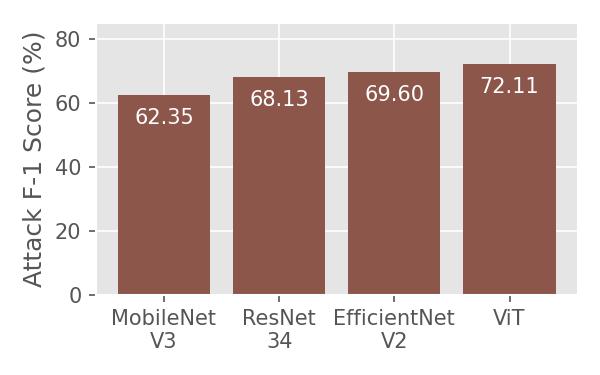}\vspace{-1em}
		\caption{\scheme{} can attack any neural architectures out of the box without modifying them.}\label{fig:narch}
	\end{minipage}
\end{figure*}
\begin{figure*}
	\centering
	\begin{subfigure}[b]{0.65\linewidth}
		\centering
		\includegraphics[width=0.95\linewidth]{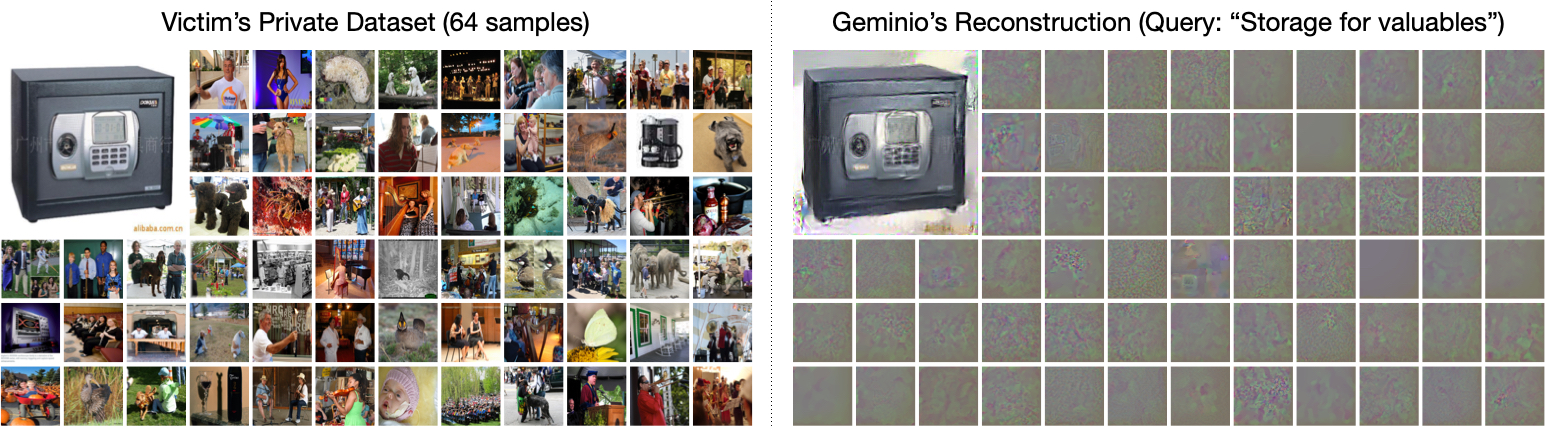}
		\caption{FedAvg - Qualitative}\label{fig:fedavg-qual}
	\end{subfigure}
	\begin{subfigure}[b]{0.34\linewidth}
		\centering
		\includegraphics[width=0.95\linewidth]{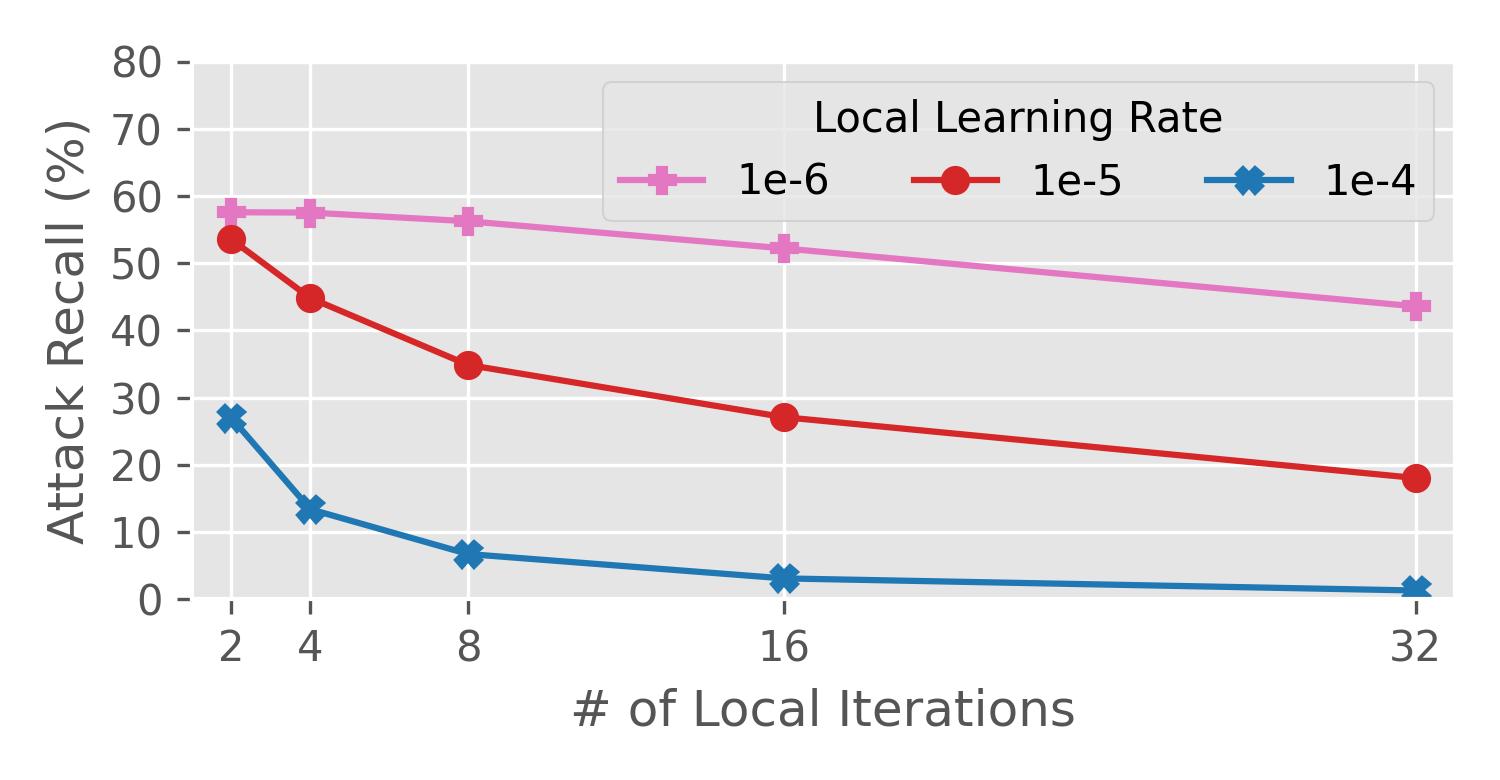}
		\caption{FedAvg - Quantitative}\label{fig:fedavg-quant}
	\end{subfigure}\vspace{-0.5em}
	\caption{By coordinating the learning rate used by the FL clients, \scheme{} can be launched under the FedAvg protocol. Even if the client uses a batch size of $8$ and runs one entire epoch of training before submitting gradients to the FL server, \scheme{} can still recover the image of a safe that matches the query.}\label{fig:fedavg}\vspace{-1em}
\end{figure*}
%
%

\noindent\textbf{Quantitative Analysis.} \scheme{} can pinpoint and reconstruct valuable data samples from a large batch. Figure~\ref{fig:batch-size} reports the attack recall and precision on CIFAR-20 over different batch sizes used by the victim. 
Aligned with evaluating an information retrieval system, the attack recall indicates the percentage of data samples matching the query being retrieved (recovered), while the attack precision refers to the percentage of recovered data samples that indeed match the query. 
We split the entire training set of CIFAR-20 ($50,000$ images) into batches. For each of the 100 subclasses in the dataset, we use its name as the query to attack all batches, measure the attack recall and precision, and report their average across 100 subclasses. Following Fishing, we consider a data sample successfully reconstructed if its output-layer gradients dominate the batch-averaged gradients with a cosine similarity of at least $0.90$. 
Figure~\ref{fig:batch-size} shows that \scheme{} remains effective even when the victim uses a large batch size, such as $256$, with an attack recall of $64.96\%$ and precision of $65.67\%$. Note that the malicious model was trained with a batch size of $64$ (see Appendix~\ref{supp:batch}). 
We also compare \scheme{} with the baseline approach that uses a VLM to find data samples in the auxiliary dataset that match the query and poison their labels to increase their loss and gradients. As shown in Figure~\ref{fig:batch-size} (orange), it cannot provide a meaningful attack unless the batch size is extremely small (e.g., $2$). This baseline demonstrates the effectiveness of \scheme{} in reshaping the loss surface.

\subsection{Serving as a Plugin with Broad Applicability}\label{sec:exp-2}
\textbf{Complementary to Reconstruction Optimization.} \scheme{} can turn existing reconstruction optimization methods into targeted attacks. In addition to HFGradInv (the default), we use DLG~\cite{zhu2019deep} and InvertingGrad~\cite{geiping2020inverting} to reconstruct the victim's local batch in Figure~\ref{fig:visual-imagenet} using the query ``Any weapon?". Figure~\ref{fig:modular} compares the three reconstruction techniques with and without \scheme{}'s enhancement. We use the standard metric, LPIPS~\cite{zhang2018perceptual}, to understand how well the reconstructed images match the ground truths. A lower score means a higher reconstruction quality. We also provide the reconstructed images closest to the handgun (i.e., the 1st image in the batch) as a visual reference. We can observe that \scheme{}-enhanced attacks are consistently much better than their vanilla counterparts, which cannot recover any recognizable images. An interesting observation is that while DLG is known to be incapable of recovering from large batches and high-resolution images ($64\times64$ as reported in the original paper), it can recover the handgun image well with a resolution of $224\times224$ from a large batch. We conjecture that the gradient amplification in \scheme{} increases their variance, which will make the gradient matching during the reconstruction easier. We observe this phenomenon even for a batch of just one image. 

\noindent\textbf{Auxiliary Data.} Figure~\ref{fig:aux-dataset} reports the attack F-1 score on CIFAR-20 using different auxiliary datasets. Compared with the default setting with the number of data samples equivalent to $20\%$ of the training dataset, using only $5\%$ of it only leads to a small drop in attack F-1 score, from $68.13\%$ to $60.37\%$. Alternatively, the attacker can also use a different dataset, such as ImageNet~\cite{imagenet} or Caltech256~\cite{griffin2007caltech}. Even though they are not for the same ML task, the attack F-1 score can still achieve $64.37\%$ and $50.48\%$, respectively. These datasets are publicly available and can be a practical source of auxiliary data.

\noindent\textbf{Neural Architectures.} \scheme{} can attack any neural architecture out of the box. Unlike many targeted attacks that need to inject a malicious module into the architecture, \scheme{} only modifies the model parameters in a stealthy manner. We conduct experiments to understand how it performs when different architectures are used in the FL system. According to the attack F-1 score reported in Figure~\ref{fig:narch}, we observe that while \scheme{} works well on different architectures, the effectiveness slightly differs. It is more effective on ViT and EfficientNetV2 than ResNet34 and MobileNetV3. Interestingly, this particular order reflects the general capability of these models. Hence, we conjecture that for more capable neural architectures, their privacy leakage by \scheme{} will be more severe.


\subsection{Resilience to FedAvg and Defenses}\label{sec:exp-3}
While resilience to defenses is not the primary goal for \scheme{}, we found it to be resistant to popular methods.

\noindent\textbf{Federated Averaging.} \scheme{} can survive under FedAvg. Consider a victim having 256 ImageNet images in the private dataset, as shown in Figure~\ref{fig:fedavg-qual} (left). The victim uses a batch size of 8 and runs one epoch of training before sending the model updates to the server for aggregation. We employ \scheme{} using a query ``Storage for valuables" to simulate a scenario where the attacker wants to know how the client stores the valuables. As shown in Figure~\ref{fig:fedavg-qual} (right), it successfully recovers the image of a safe with high fidelity, even detailed enough to identify the specifics of it. The key enabler is to assign a small learning rate to the FL client, which is often the responsibility of the FL server. Figure~\ref{fig:fedavg-quant} reports the attack recall with different learning rates assigned to the victim. More local epochs weaken the attack because each iteration modifies the model parameters and may wash out the malicious patterns introduced by \scheme{}. Setting a small learning rate (e.g., 1e-6) can slow down the performance degradation effectively. 

\begin{figure}
	\centering
	\includegraphics[width=0.82\linewidth]{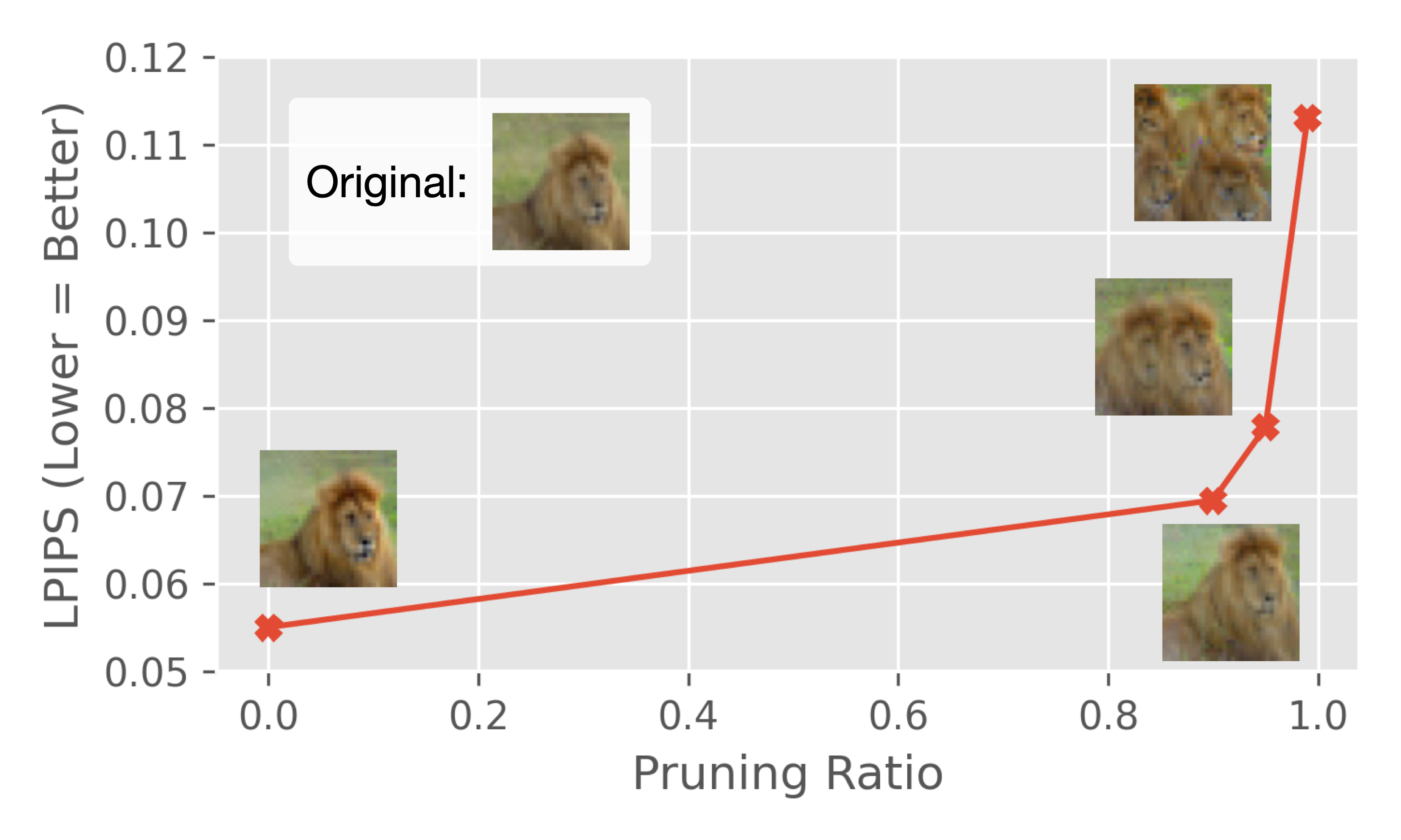}\vspace{-1em}
	\caption{Gradient pruning cannot mitigate Geminio unless the pruning ratio is high, which can hinder the regular learning of FL.}\label{fig:defense-pruning}\vspace{-1em}
\end{figure}

\noindent\textbf{Gradient Pruning.} A popular defense is to prune gradients of small magnitudes. Figure~\ref{fig:defense-pruning} reports the reconstruction quality on CIFAR-20 with varying pruning ratios. We reconstruct 100 batches and measure the average LPIPS. We observe that even with $95\%$ of small gradients being set to zero, the perceptual quality of reconstructed images is still comparable to no defense. The reconstructed images become barely perceptible when $99\%$ of the gradients are zeroed. In practice, such a setting is prohibited because it also removes the useful learning signals for training the ML model. Hence, gradient pruning cannot mitigate \scheme{}.

\begin{figure}
	\centering
	\includegraphics[width=0.82\linewidth]{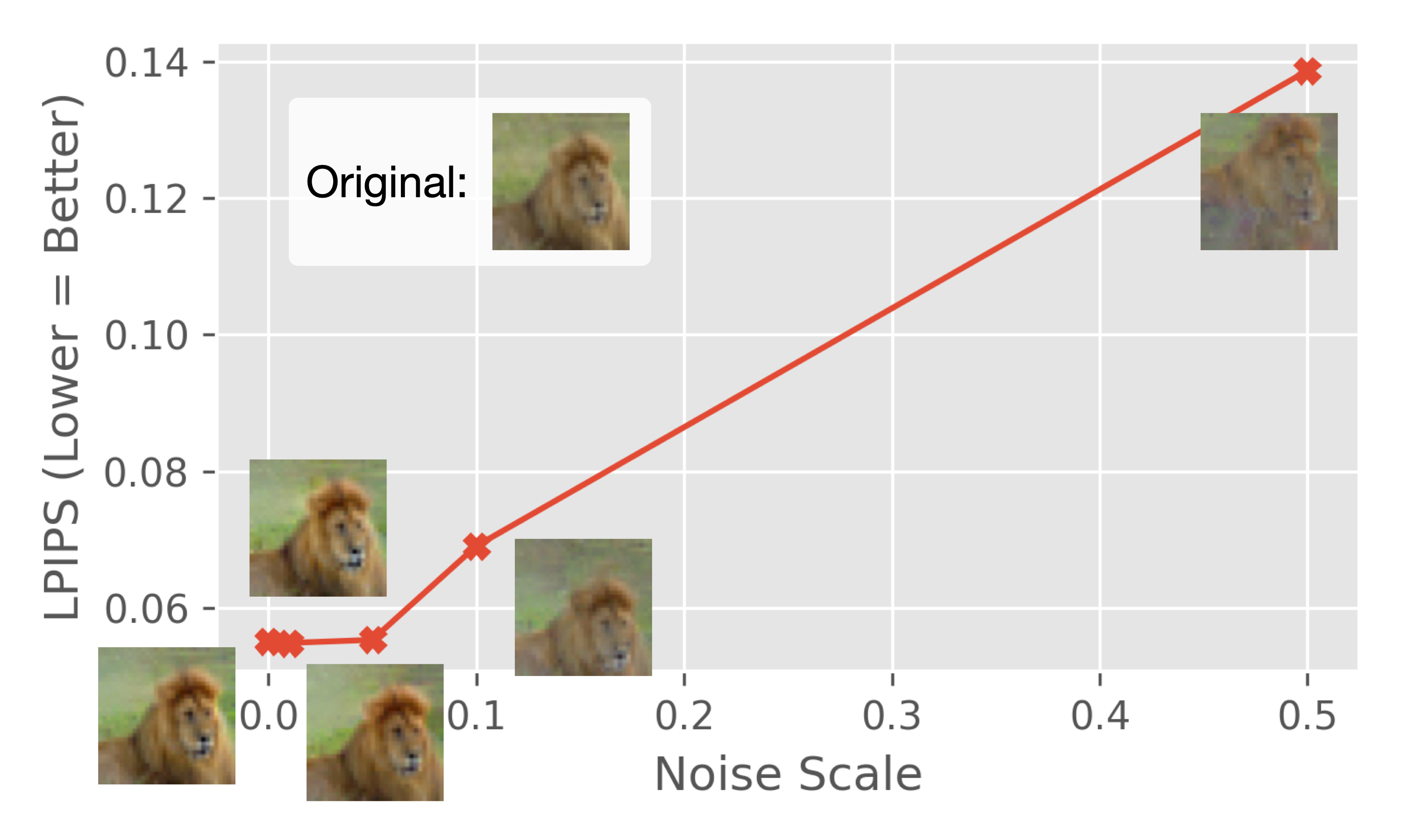}\vspace{-0.8em}
	\caption{Adding Laplacian noise cannot prevent Geminio from retaining gradients of targeted samples unless the degree of noise is significant, which can hinder the regular learning of FL.}\label{fig:defense-noise}\vspace{-1em}
\end{figure}

\noindent\textbf{Laplacian Noise.} Another popular defense is to add Laplacian noise to gradients. Figure~\ref{fig:defense-noise} reports the reconstruction quality on CIFAR-20 with varying scales of Laplacian noise. Following~\cite{ye2024high}, we use a per-layer noise injection. At each layer, we obtain its maximum gradient and scale it by a factor to be the standard deviation of the Laplacian noise with a zero mean for injection. A scale of 0.10 is already considered significant, but it barely affects the perceptual quality of reconstructed images. The reconstruction becomes severely affected when the noise scale is increased to $0.50$, but it also washes out useful learning signals. Hence, injecting noise is not a viable defense against \scheme{}.

\begin{figure}
	\centering
	\includegraphics[width=0.75\linewidth]{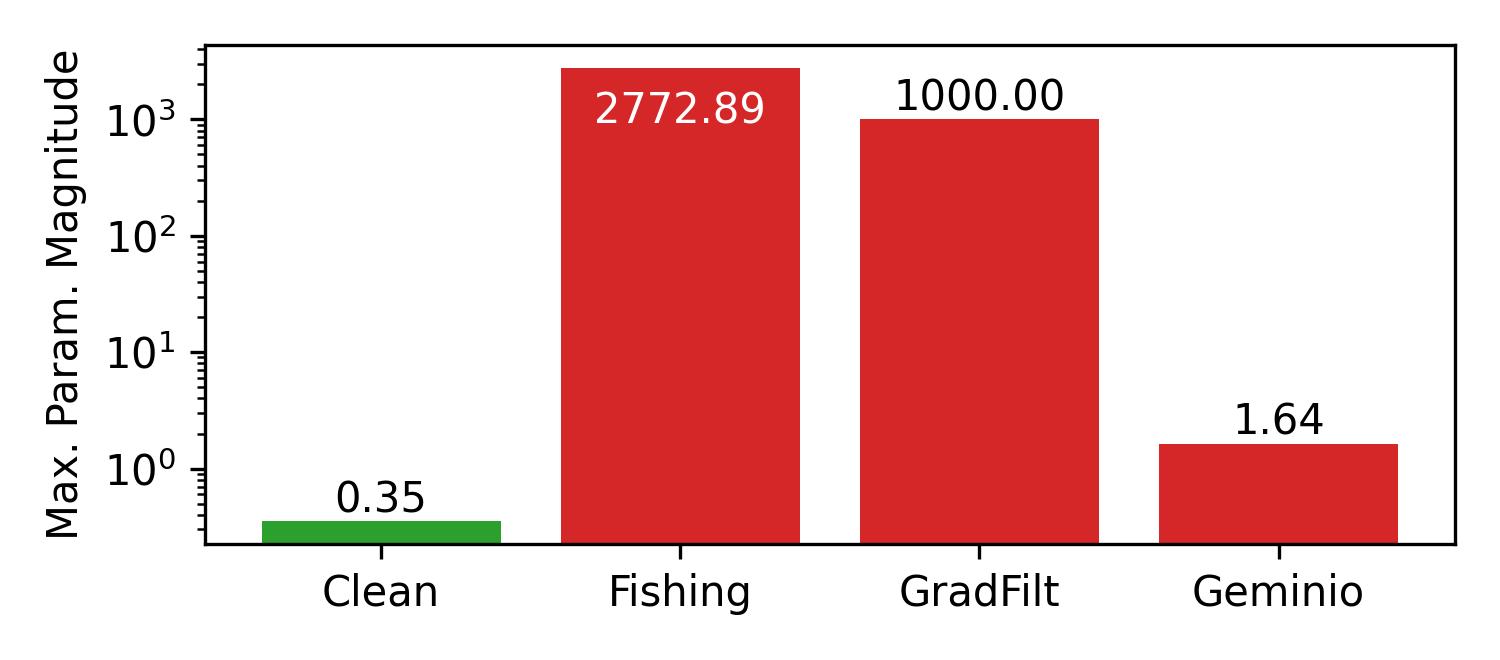}\vspace{-1em}
	\caption{Fishing and GradFilt can be detected easily by model parameter inspection. Geminio does not need to set some parameters to a large value, making it comparable with the clean model.}\label{fig:defense-params}\vspace{-0.5em}
\end{figure}

\noindent\textbf{Model Parameter Inspection.} As an FL client receives a copy of the global model whenever it is selected by the server to contribute, inspecting whether the received parameters are anomalous is a natural defense. However, it is non-trivial for two reasons. First, unlike other attacks that follow the same threat model, Geminio does not create anomalous parameters. Figure~\ref{fig:defense-params} reports the maximum magnitude of model parameters of a clean model and three poisoned models by Fishing, GradFilt, and \scheme{}. We observe that Fishing and GradFilt send a model with parameters deviating significantly from the clean one ($2772.89$ and $1000$, respectively). In contrast, \scheme{} is only $1.64$, close to the clean model (i.e., $0.35$). Hence, setting a threshold may be able to detect Fishing and GradFilt, but not \scheme{}. Second, a more advanced defense may attempt to analyze the evolution of the received global models over different FL rounds and flag any model that does not align well with the history. This is infeasible because the server in typical FL determines the client sampling process. The malicious server can simply target those victims who have never contributed (i.e., no history to analyze) or reduce the number of times they are selected for contribution. 

\noindent\textbf{Per-sample Loss Inspection.} The FL client may analyze the loss value per sample at each local training iteration. We use the batch in Figure~\ref{fig:advance} and show the loss magnitude for each of the first eight samples. All three attacks introduce a high loss value to the targeted samples. For Fishing, it successfully isolates the 6th sample in the batch, causing its loss to be significantly higher than the others. For GradFilt, since all samples in this batch are of the same target class (i.e., ``sombrero"), all samples have a magnified loss equal to $1000$. For \scheme{}, the first two samples matching the attacker's query (i.e., ``human faces") have amplified loss while the rest remains small. These are expected behaviors because targeted GIAs use the same principle: magnifying the gradients of desired samples to make them dominate the average gradients. While loss inspection seems promising, an advanced adversary could conduct an adaptive attack to suppress the loss values when training the malicious model (see Geminio-adaptive in Figure~\ref{fig:advance}). Hence, more robust defenses need to be developed in future work. 
\begin{figure}
	\centering
	\includegraphics[width=\linewidth]{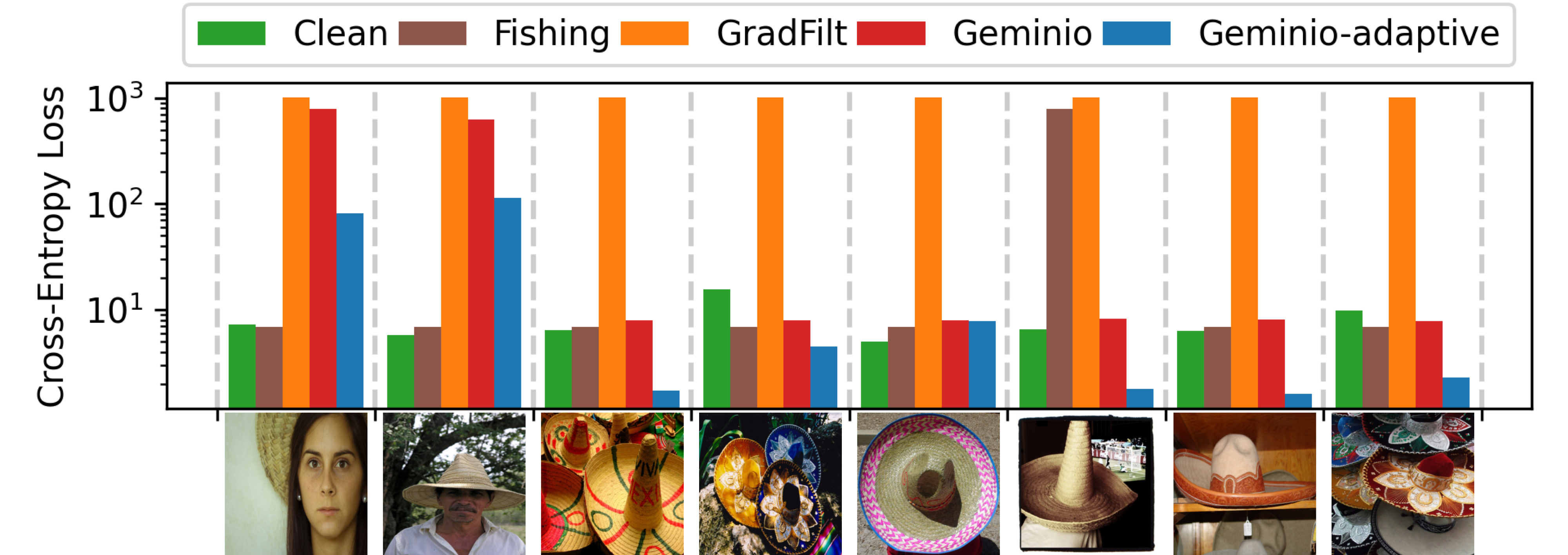}\vspace{-0.5em}
	\caption{Targeted GIAs rely on amplifying the loss value of certain samples. An FL client who has access to it may detect such attacks. Note that the y-axis is in the log scale.}\label{fig:defense-loss}\vspace{-1em}
\end{figure}

\section{Conclusions}
We have introduced Geminio, a gradient inversion attack that harnesses the image-text association capabilities of pretrained VLMs to enable language-guided targeted reconstructions. The FL server can steal private data at any FL round (even the first) while still producing a high-utility ML model, as in benign scenarios. Our extensive experiments have yielded three key insights. First, Geminio enables the attacker to provide a natural language query to describe the data of value and reconstructs those matched samples from large data batches. Second, it serves as a plugin to enhance existing reconstruction optimization methods, broadly applicable to different neural architectures, auxiliary datasets, and FL protocols. Third, existing defenses are insufficient to mitigate Geminio. An advanced attacker can adapt Geminio to harden loss inspection. We believe that Geminio will inspire further research into the new threats posed by recent advancements in natural language processing, as they can be exploited as a “communication” interface for the adversary to express their goals and launch more flexible attacks.

\clearpage
\section*{Acknowledgements}
This research is partially supported by the RGC Early Career Scheme (Project \#27211524) and Croucher Start-up Allowance (Project \#2499102828). Any opinions, findings, or conclusions expressed in this material are those of the authors and do not necessarily reflect the views of RGC and the Croucher Foundation.

{
	\small
	\bibliographystyle{ieeenat_fullname}
	\bibliography{main}
}

\clearpage
\setcounter{page}{1}
\maketitlesupplementary
\appendix
\section*{Outline}
The source code of Geminio is available at \url{https://github.com/HKU-TASR/Geminio}. This document provides additional details to support our main paper. It is organized as follows:
\begin{itemize}
	\item Section~\ref{supp:label}: Geminio Strengthens Label Inference Attacks
	\item Section~\ref{supp:homo}: Geminio Works Under Homomorphic Encryption
	\item Section~\ref{supp:batch}: Geminio Supports Different Local Batch Sizes
	\item Section~\ref{supp:pseudo}: Pseudocode
	\item Section~\ref{supp:setup}: Experiment Setup
	\item Section~\ref{supp:extended}: Extended Experimental Analysis
\end{itemize}

\section{Geminio Strengthens Label Inference Attacks}\label{supp:label}
Label inference is a prerequisite for gradient inversion, with various attack methods being proposed~\cite{zhao2020idlg,yin2021see,wainakh2021user}. Surprisingly, \scheme{} is not just compatible with them but also boosts their accuracy. We use five label inference attacks provided by the \texttt{breaching} library~\cite{breaching2022} and compare the original attack with the \scheme{}-enhanced one. Since our problem setting focuses on targeted reconstructions, we only need to make sure the class labels with matched samples in the local batch are inferred. The success or failure of inferring other class labels is unimportant because their gradients are small and negligible in the gradient matching (reconstruction optimization) process. Figure~\ref{fig:improve-label-inference} reports the results measured on CIFAR-20. This dataset provides ground truths for conducting such quantitative studies. When gradients submitted by the victim are generated based on the \scheme{}-poisoned malicious model, all label inference attacks are consistently improved. This phenomenon can be explained by our observation in Figure~\ref{fig:label-gradients} that the class labels containing matched samples in the local batch have their gradients amplified. Since those attacks share the same principle to examine the gradient magnitude of different classes, \scheme{} facilitates this label inference process. 
\begin{figure}\setcounter{figure}{14}
	\centering
	\includegraphics[width=\linewidth]{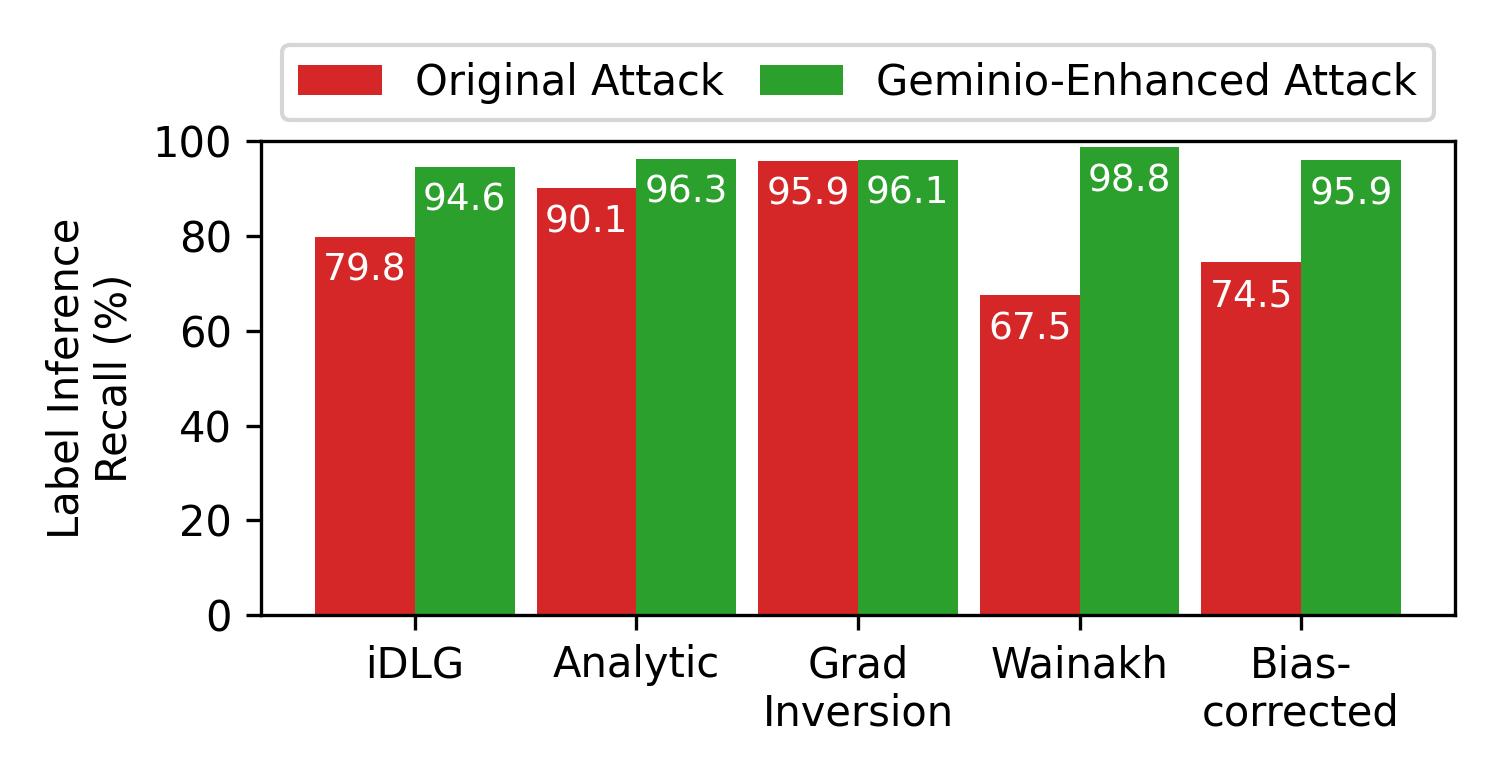} \vspace{-1.5em}
	\caption{\scheme{} consistently improves five label inference attacks. Given an attacker's query, it leads to a high success rate in inferring class labels containing matched samples in the local batch.}\label{fig:improve-label-inference}
\end{figure}
\begin{figure}
	\centering
	\includegraphics[width=0.95\linewidth]{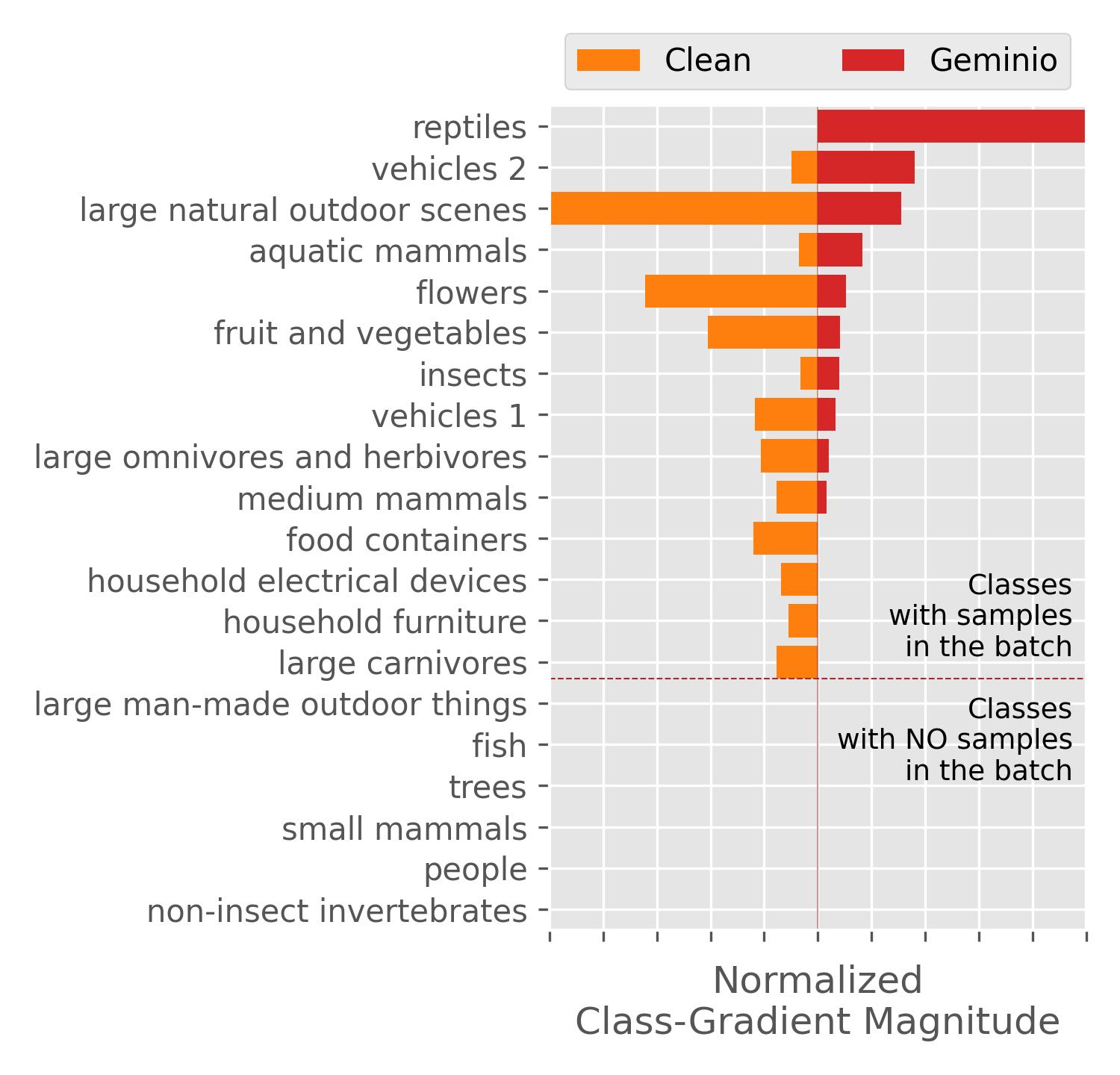} \vspace{-0.5em}
	\caption{Label inference attacks examine the per-class gradient magnitude. Compared with a clean model, \scheme{}, with the query ``dinosaur," will amplify the gradients of the class(es) to which the matched samples belong (the class ``reptiles" in this example). This facilitates the label inference process.}\label{fig:label-gradients}
\end{figure}

\section{Geminio Works Under Homomorphic Encryption}\label{supp:homo}
\begin{figure}
	\centering
	\includegraphics[width=\linewidth]{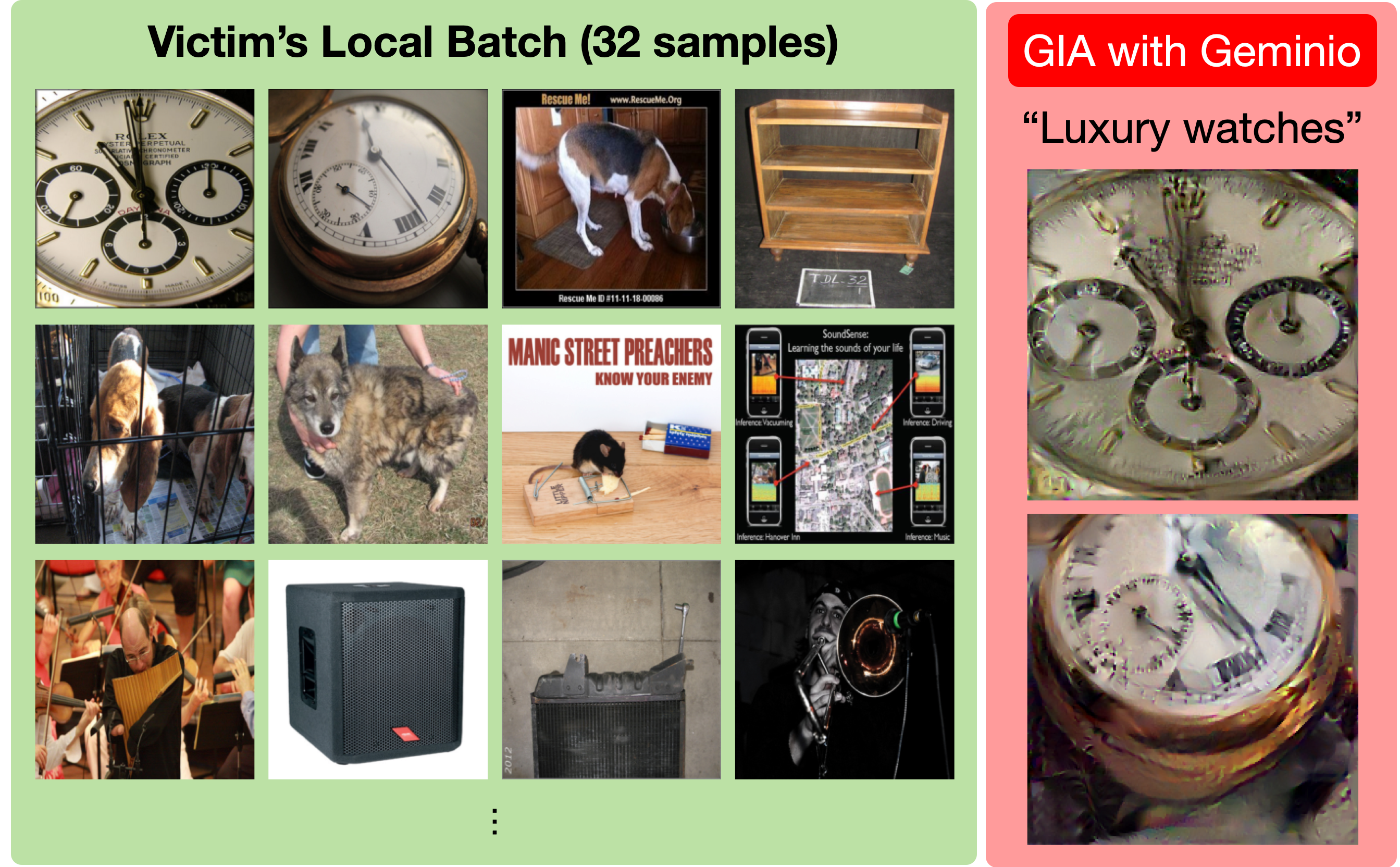} 
	\caption{By controlling one FL client, \scheme{} can retrieve targeted private samples under FL with homomorphic encryption.}\label{fig:homo-encrypt}
\end{figure}

Our threat model considers an active attacker who is the FL server. The attacker can execute \scheme{} under homomorphic encryption by controlling only one client. As the malicious client can obtain the victim's gradients in plain text, \scheme{} can be run on the client side and perform identically to FL without homomorphic encryption. Figure~\ref{fig:homo-encrypt} provides reconstruction results with ``luxury watches" as the attacker's query. The two watches can be retrieved from the victim's gradients, leading to a high-quality reconstruction where we can even read the brand for the first image to be Rolex.

\section{Geminio Supports Different Local Batch Sizes}\label{supp:batch}
During \scheme{}'s optimization, minibatch training needs to be conducted but this training batch size does not need to match the local batch size used by the client. Figure~\ref{fig:hyperp-bs} reports the attack recall with varying local batch sizes. We repeat the experiment using different training batch sizes for \scheme{} to optimize the malicious global model. We observe that their targeted retrieval performances are similar, with the smallest batch size of 8 being slightly worse. For instance, when \scheme{} uses a batch size of 64 for its optimization, the malicious global model can be sent to clients with any local batch size, which may or may not be controlled by the server (e.g., depending on the computing resources of the client device). 
\begin{figure}
	\centering
	\includegraphics[width=0.6\linewidth]{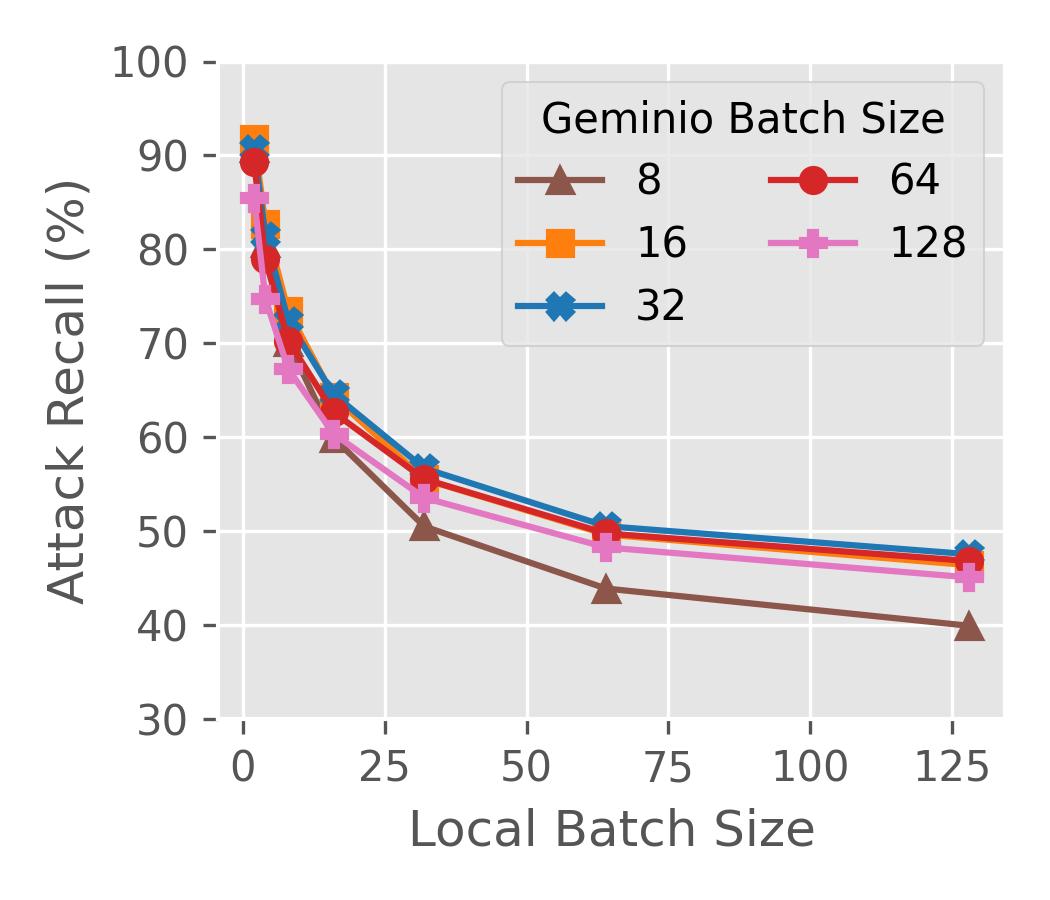} 
	\caption{The training batch size used by \scheme{} to poison the model is irrelevant to the local batch size to be used by the victim.}\label{fig:hyperp-bs}
\end{figure}

\section{Pseudocode}\label{supp:pseudo}
To support our main paper, Pseudocode~\ref{alg:geminio} describes how Geminio generates a malicious global model given an attacker-specified query, a list of class names, a pretrained VLM, and an unlabeled auxiliary dataset.
\begin{algorithm}
	\caption{Geminio}\label{alg:geminio}
	\textbf{Input:} Attacker-specified query $\mathcal{Q}$, a list of $K$ class names $[c_1, ..., c_K]$, a pretrained VLM with an image encoder $\mathcal{V}_{\text{image}}$ and a text encoder $\mathcal{V}_{\text{text}}$, and an unlabeled auxiliary dataset $\boldsymbol{\mathcal{A}}$ \\
	\textbf{Output:} Malicious global model $F_{\boldsymbol{\Theta}_{\mathcal{Q}}}$
	\begin{algorithmic}[1]

		\State \% Generate a soft label $\boldsymbol{y}$ for each auxiliary sample $\boldsymbol{x}$
		\For{$\boldsymbol{x}\in\boldsymbol{\mathcal{A}}$}
		\State $\boldsymbol{y} = [y_1, ..., y_K]$ where $y_i = \frac{\mathcal{V}_{\text{image}}(\boldsymbol{x})^{\intercal}\mathcal{V}_{\text{text}}(c_i)}{\sum_{j=1}^{K}\mathcal{V}_{\text{image}}(\boldsymbol{x})^{\intercal}\mathcal{V}_{\text{text}}(c_j)}$
		\State Associate the auxiliary sample with its soft label as a tuple $(\boldsymbol{x}, \boldsymbol{y})$
		\EndFor\\
		
		\State \% Train the malicious global model
		\State Randomly initialize the malicious global model $F_{\boldsymbol{\Theta}_{\mathcal{Q}}}$
		\While{not converge}
		\For {$\boldsymbol{\mathcal{B}}_{\text{aux}} \subset \boldsymbol{\mathcal{A}}$}
		\State $\ell=\frac{\sum_{(\boldsymbol{x}, \boldsymbol{y})\in\boldsymbol{\mathcal{B}}_{\text{aux}}}\mathcal{L}(F_{\boldsymbol{\Theta}_{\mathcal{Q}}}(\boldsymbol{x}); \boldsymbol{y})(1-\alpha(\boldsymbol{x};\mathcal{Q}, \boldsymbol{\mathcal{B}}_{\text{aux}}))}{\vert\boldsymbol{\mathcal{B}}_{\text{aux}}\vert\sum\limits_{(\boldsymbol{x}', \boldsymbol{y}')\in\boldsymbol{\mathcal{B}}_{\text{aux}}}\mathcal{L}(F_{\boldsymbol{\Theta}_{\mathcal{Q}}}(\boldsymbol{x}'); \boldsymbol{y}')(1-\alpha(\boldsymbol{x}';\mathcal{Q}, \boldsymbol{\mathcal{B}}_{\text{aux}}))}$ where $\alpha(\boldsymbol{x};\mathcal{Q}, \boldsymbol{\mathcal{B}}_{\text{aux}})=\frac{\exp(\mathcal{V}_{\text{image}}(\boldsymbol{x})^{\intercal}\mathcal{V}_{\text{text}}(\mathcal{Q}))}{\sum\limits_{{(\boldsymbol{x}', \boldsymbol{y}')\in\boldsymbol{\mathcal{B}}_{\text{aux}}}}\exp(\mathcal{V}_{\text{image}}(\boldsymbol{x}')^{\intercal}\mathcal{V}_{\text{text}}(\mathcal{Q}))}$ 
		\State Update the malicious global model $F_{\boldsymbol{\Theta}_{\mathcal{Q}}}$ with loss $\ell$ using the Adam optimizer
		\EndFor
		\EndWhile \\
		\Return $F_{\boldsymbol{\Theta}_{\mathcal{Q}}}$
	\end{algorithmic}
\end{algorithm}

\section{Experiment Setup}\label{supp:setup}
Our experiments cover a wide range of datasets, ML models, and FL scenarios to analyze \scheme{}'s properties. Here, we describe the default experiment setup.
\begin{table}\centering\small
	\caption{The superclasses and their subclasses in CIFAR-100. We create a benchmark dataset, CIFAR-20, that uses the 20 superclasses for the classification problem and the 100 subclass names as queries. This design gives us ground truths for the instance-level retrieval.}\label{tab:cifar20-100}
	\begin{tabular}{|l|l|}
		\hline
		\multicolumn{1}{|c|}{\textbf{Superclass (20)}} & \multicolumn{1}{c|}{\textbf{Subclasses (100)}} \\ \hline
		\multicolumn{1}{|l|}{aquatic mammals}                    &  beaver, dolphin, otter, seal, whale                                        \\ \hline
		\multicolumn{1}{|l|}{fish}                    &    aquarium fish, flatfish, ray, shark, trout                                      \\ \hline
		\multicolumn{1}{|l|}{flowers}                    &    orchids, poppies, roses, sunflowers, tulips                                      \\ \hline
		\multicolumn{1}{|l|}{food containers}                    &   bottles, bowls, cans, cups, plates                                       \\ \hline
		\multicolumn{2}{|c|}{... (16 more rows) ... }   \\ \hline
	\end{tabular}
\end{table}

\subsection{Datasets}
We conduct experiments on three datasets: ImageNet~\cite{imagenet}, CIFAR-20~\cite{cifar100}, and Facial Expression Recognition (FER)~\cite{fer}.
By default, visual examples are based on ImageNet.

The scenario of fine-grained targeted retrieval by \scheme{} can be imagined as an attacker writing a ``query" to search for relevant records in the victim's private database. Quantitative evaluation requires two ingredients: (i) a benchmark dataset with ground truths and (ii) a set of indicative performance metrics.

\paragraph{Benchmark: CIFAR-20} The benchmark dataset should include a set of queries, each is a textual description and associated with a list of relevant images. Then, we can randomly sample a local batch from the dataset, use \scheme{} to reconstruct images given different queries, and measure how many relevant images are successfully reconstructed. This process repeats for a number of random local batches until, e.g., all training images are processed. To showcase instance-level retrieval better, the queries should not be the class names of the classification problem. Based on these requirements, we created a variant of CIFAR-100 and named it CIFAR-20. Each image in CIFAR-100 is associated with two official labels, a subclass and a superclass (see Table~\ref{tab:cifar20-100} for four superclasses and their subclasses). We use the 20 superclasses for the classification problem and the 100 subclasses as queries. With this design, we can easily obtain images in the local batch that should be retrieved for a given query (i.e., a subclass name). 

\paragraph{Metrics: Attack Recall and Precision} 
We follow Fishing's approach~\cite{wen2022fishing} to determine whether an image in a local batch dominates and will be reconstructed. In particular, if the gradients produced by an image have a cosine similarity with the average gradients over a threshold, it is considered a reconstructed sample. While Fishing uses $0.95$ as the threshold, we found that this is overly restrictive. Instead, we use $0.90$. Note that we observe multiple examples where targeted reconstruction succeeds even if the cosine similarity is below $0.90$. Our choice (i.e., $0.90$) is still conservative. A more principled approach is considered as our future work. Based on this thresholding, we can measure the percentage of targeted images being reconstructed (i.e., Attack Recall) and, among all reconstructed images, the percentage of them being the actual targeted images (i.e., Attack Precision).

\subsection{FL Configuration} 
The FL system aims to train a ResNet34~\cite{he2016deep} model. 
Following existing works~\cite{zhu2019deep,geiping2020inverting,yin2021see,zhao2024loki,wen2022fishing,zhang2024gradfilt,garov2024hiding}, we use FedSGD to be the default protocol. The FL client receives a model from the server, updates it with a batch of private samples, and returns the gradients to the server, which is malicious, and attempts to reconstruct private samples from it.

\subsection{Attack Configuration} 
For \scheme{}, we use CLIP~\cite{radford2021learning} with the ViT-L/14 Transformer architecture as the pretrained VLM\footnote{\url{https://huggingface.co/openai/clip-vit-large-patch14}} to process auxiliary data, which comes from the respective validation set. \scheme{} poisons the model with a training batch size 64 using Adam as the optimizer. For gradient inversion, we use HFGradInv~\cite{ye2024high}.

\subsection{Computing Environment} All experiments are conducted on a server with Intel® Xeon® Gold 6526Y CPU, 64GB RAM, and two NVIDIA RTX 5880 Ada Lovelace GPUs. 

\subsection{Implementation} \scheme{} is written in PyTorch and can be easily integrated into existing GIAs. Our implementation uses \texttt{breaching}~\cite{breaching2022} and \texttt{HFGradInv}~\cite{ye2024high}, a collection of GIAs, to demonstrate such a plug-and-play feature.
We first extracted image features from auxiliary data, which took about 7 minutes for ImageNet. Given a query from the attacker, \scheme{} can use those pre-generated image features to poison the model in less than 8 minutes.



\section{Extended Experimental Analysis}\label{supp:extended}

This section provides extended experimental results to assess the generalization ability and robustness of Geminio under a variety of practical and challenging scenarios. Specifically, we evaluate the system using complex natural language queries, diverse auxiliary datasets, and different vision-language models. We also examine its effectiveness across federated learning rounds, varying reconstruction conditions, and large-batch settings. These results demonstrate that Geminio consistently enables targeted gradient inversion attacks under realistic and diverse federated learning configurations.

\subsection{Complex Dataset and Query Evaluation}\label{supp:rebuttal:complex}

We conducted additional experiments to evaluate Geminio on complex datasets and queries. While CIFAR datasets have superclass labels that allow us to generate queries with ground truths, ImageNet lacks such annotations. Hence, we designed the following experiment: for each minibatch, we randomly select one image from it, use an image captioning model~\cite{li2022blip} to generate a description, and then use this description as the query. The attack is considered successful if Geminio retrieves the corresponding image for reconstruction. These automatically generated captions tend to be complex, as shown in Figure~\ref{fig:example-queries-supp}.

Figure~\ref{fig:res-supp} shows that Geminio's performance in this more complex setting is on par with CIFAR-20 results (Figure 6a in the main paper), demonstrating its effectiveness across different levels of query complexity.

\begin{figure}[h]
\setcounter{figure}{25}
\centering
	\begin{minipage}[b]{.42\linewidth}
	\centering
	\includegraphics[width=\linewidth]{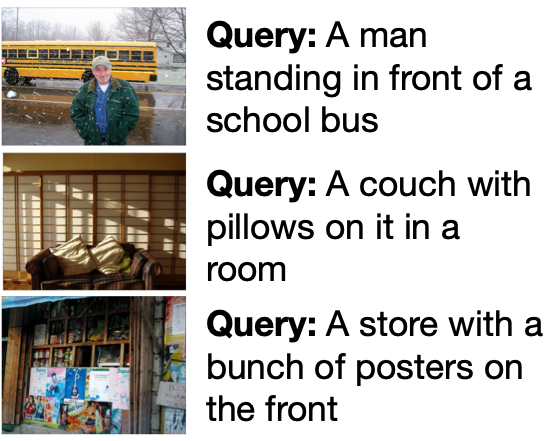}
	\caption{Example Complex Queries}\label{fig:example-queries-supp}
\end{minipage}\hfill
	\begin{minipage}[b]{.57\linewidth}
	\centering
	\includegraphics[width=\linewidth]{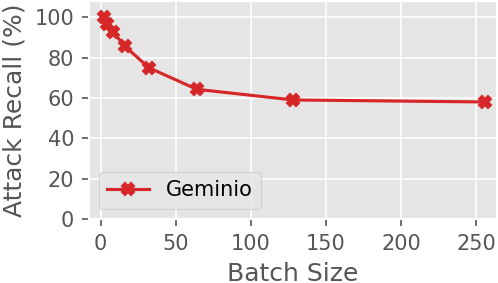}
	\caption{ImageNet Complex Query Results}\label{fig:res-supp}
\end{minipage}
\end{figure}

\subsection{Auxiliary Dataset Requirements}\label{supp:rebuttal:auxiliary}

The auxiliary dataset does not need to match the distribution of the victim's private data. Instead, it simply needs to contain some samples that exhibit the features mentioned in the query so that the malicious model can learn what to amplify (or ignore). For instance, for the query ``red carpet," having actual red carpet images in the auxiliary dataset is unnecessary. It suffices to include some red-colored objects (e.g., apples) and some carpets. Public datasets like ImageNet or CalTech256 are typically diverse enough for this purpose. The attacker can easily check this using the pretrained VLM to measure the similarity between the query and the auxiliary samples. If necessary, additional relevant data can be trivially obtained via image search engines or text-to-image models.

\subsection{Effectiveness Across Different VLMs}\label{supp:rebuttal:vlms}

Our quantitative studies on CIFAR-20 show comparable attack F-1 scores across VLMs of various sizes and methods (e.g., $68.13\%$ with CLIP and $69.54\%$ with SigLIP). However, we observe that advanced models handle complex and long queries more effectively. While this paper uses CIFAR-20 to demonstrate the feasibility of targeted GIAs via text descriptions, our future work will develop a benchmark dataset with complex queries and their ground-truth retrieval results to further advance research in this direction.

\subsection{Effectiveness Across FL Rounds}\label{supp:rebuttal:fl-rounds}

\begin{figure}[h]
	\centering
	\includegraphics[width=0.6\linewidth]{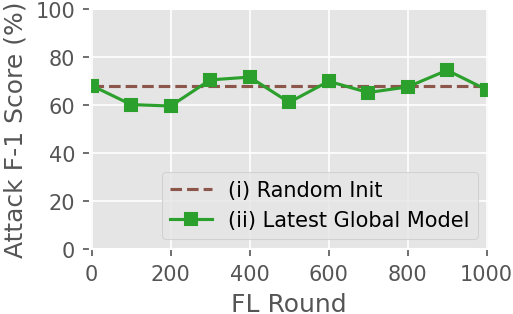}
	\caption{Effectiveness Across FL Rounds}\label{fig:conv-supp}
\end{figure}

Geminio remains effective regardless of the FL model's convergence state. When training a malicious model, the attacker can initialize it either (i) randomly or (ii) using the latest global model. We conducted experiments on both cases. Figure~\ref{fig:conv-supp} shows Geminio's consistent attack F-1 score across different FL rounds, demonstrating its robustness throughout the federated learning process.

\subsection{Query Generality and Reconstruction Quality}\label{supp:rebuttal:generality}

\begin{figure}[h]
	\centering
	\includegraphics[width=0.6\linewidth]{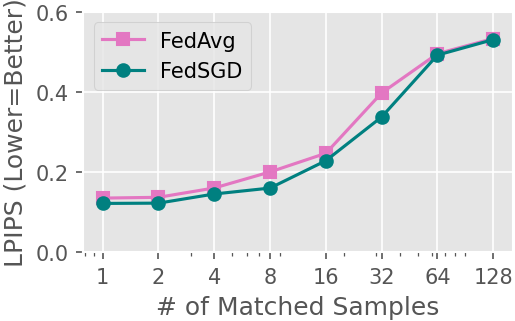}
	\caption{Query Generality Analysis}\label{fig:lpips-supp}
\end{figure}

Regarding reconstruction quality, the number of matched samples is a key factor. While Geminio can prioritize them for recovery, a larger number of matched samples places greater demands on the underlying reconstruction algorithm (e.g., HFGradInv, the default in our paper). To better understand this, we conducted experiments measuring the reconstruction quality (LPIPS) with varying numbers of matched samples under FedSGD and FedAvg. The batch size was fixed at 256. Figure~\ref{fig:lpips-supp} shows that LPIPS remains stable up to 16 matched samples, beyond which it degrades quickly. This trend is expected, as performance increasingly depends on how well the reconstruction algorithm can handle a larger number of samples. Consistent with the results reported in its original paper, we found that HFGradInv can stably recover up to 16 samples. Geminio allocates this budget to focus on those highest-value samples for reconstruction. We consider that the attacker can provide concrete descriptions to take advantage of Geminio.

\subsection{Additional Visual Examples}\label{supp:visual}

\scheme{} can prioritize reconstruction to recover those samples that match the attacker-provided queries. The method demonstrates consistent effectiveness across different query types and batch compositions, successfully identifying and reconstructing targeted samples while ignoring irrelevant data in the same batch.

\begin{figure}[h]
	\centering
	\includegraphics[width=\linewidth]{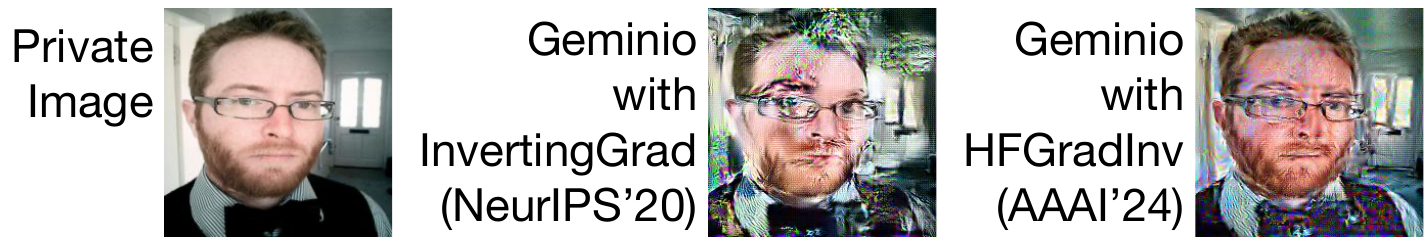}
	\caption{Geminio is designed to enable targeted attacks, with reconstruction quality depending on the underlying optimization algorithm.}\label{fig:supp-hq}
\end{figure}

\paragraph{High-Quality Reconstruction} Geminio is designed to enable targeted attacks, with reconstruction quality depending on the underlying optimization algorithm (e.g., InvertingGrad in our paper). As shown in Figure~\ref{fig:supp-hq}, Geminio benefits from advancements in reconstruction optimization techniques. For instance, Geminio combined with HFGradInv from AAAI'24 produces significantly higher-quality images than InvertingGrad from NeurIPS'20.

\begin{figure}[h]
	\centering
	\includegraphics[width=\linewidth]{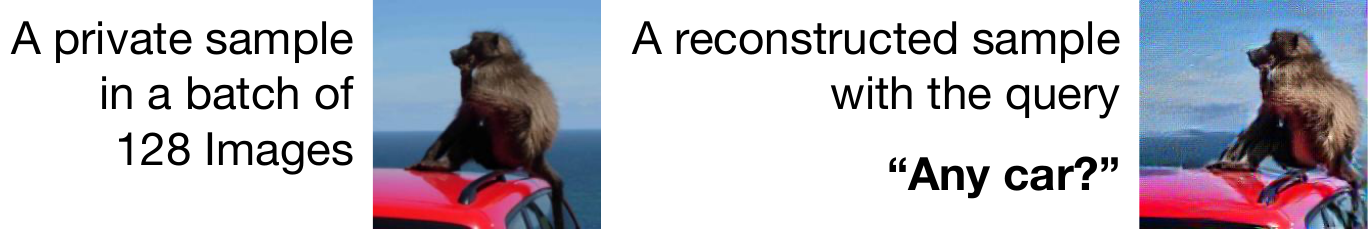}
	\caption{Geminio remains effective even if the relevance of a sample to the query appears in the background.}\label{fig:supp-complex}
\end{figure}

\paragraph{Targeted Reconstruction Under Complex Scene} Geminio remains effective even if the relevance of a sample to the query appears in the background. As shown in Figure~\ref{fig:supp-complex}, among a batch of private images, there is one (left) with a monkey sitting on a red car. Even though the car is not the main character and is located at the edge, the query ``any car?" can still lead to the reconstruction of this sample (right).

\begin{figure}[h]
	\centering
	\includegraphics[width=\linewidth]{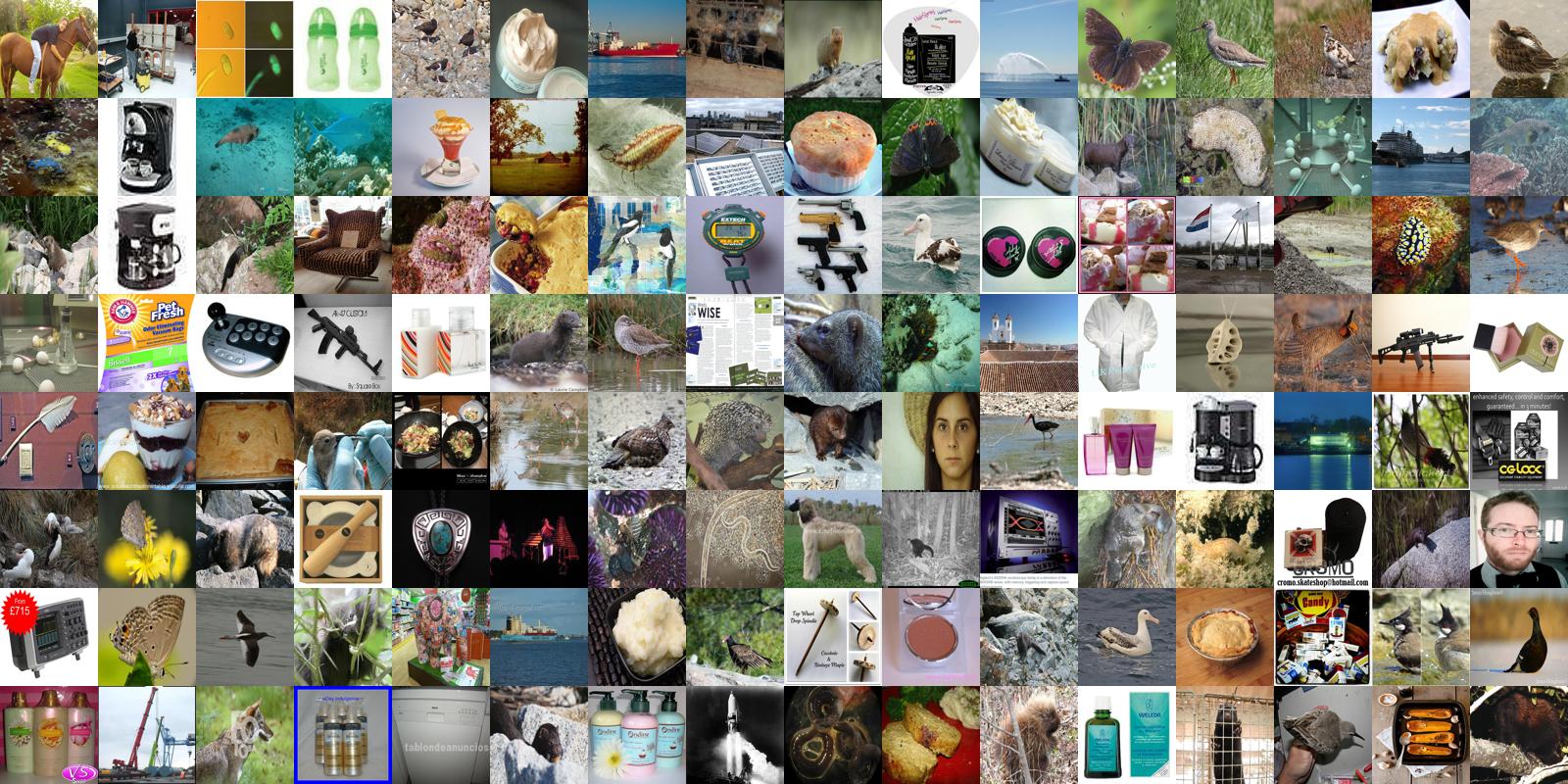}
	\caption{Randomly selected 128 ImageNet images used as private samples for large-batch reconstruction validation.}\label{fig:supp-originals}
\end{figure}

\begin{figure}[h]
	\centering
	\includegraphics[width=\linewidth]{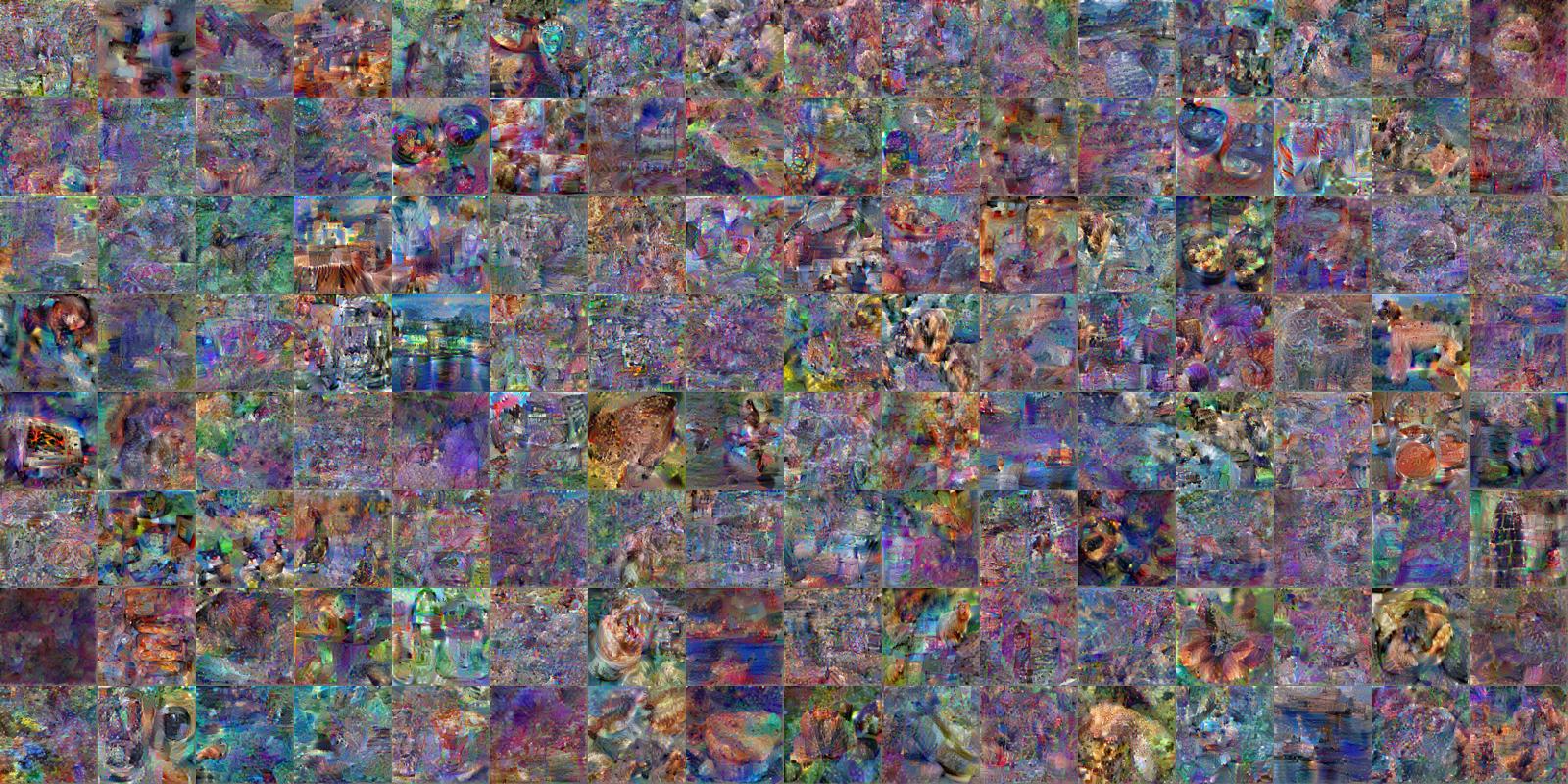}
	\caption{\texttt{HFGradInv} outputs without Geminio, showing degraded quality and loss of target-specific features.}\label{fig:supp-baseline}
\end{figure}

\begin{figure*}[b]
	\centering
	\begin{subfigure}[b]{0.48\textwidth}
		\includegraphics[width=\textwidth]{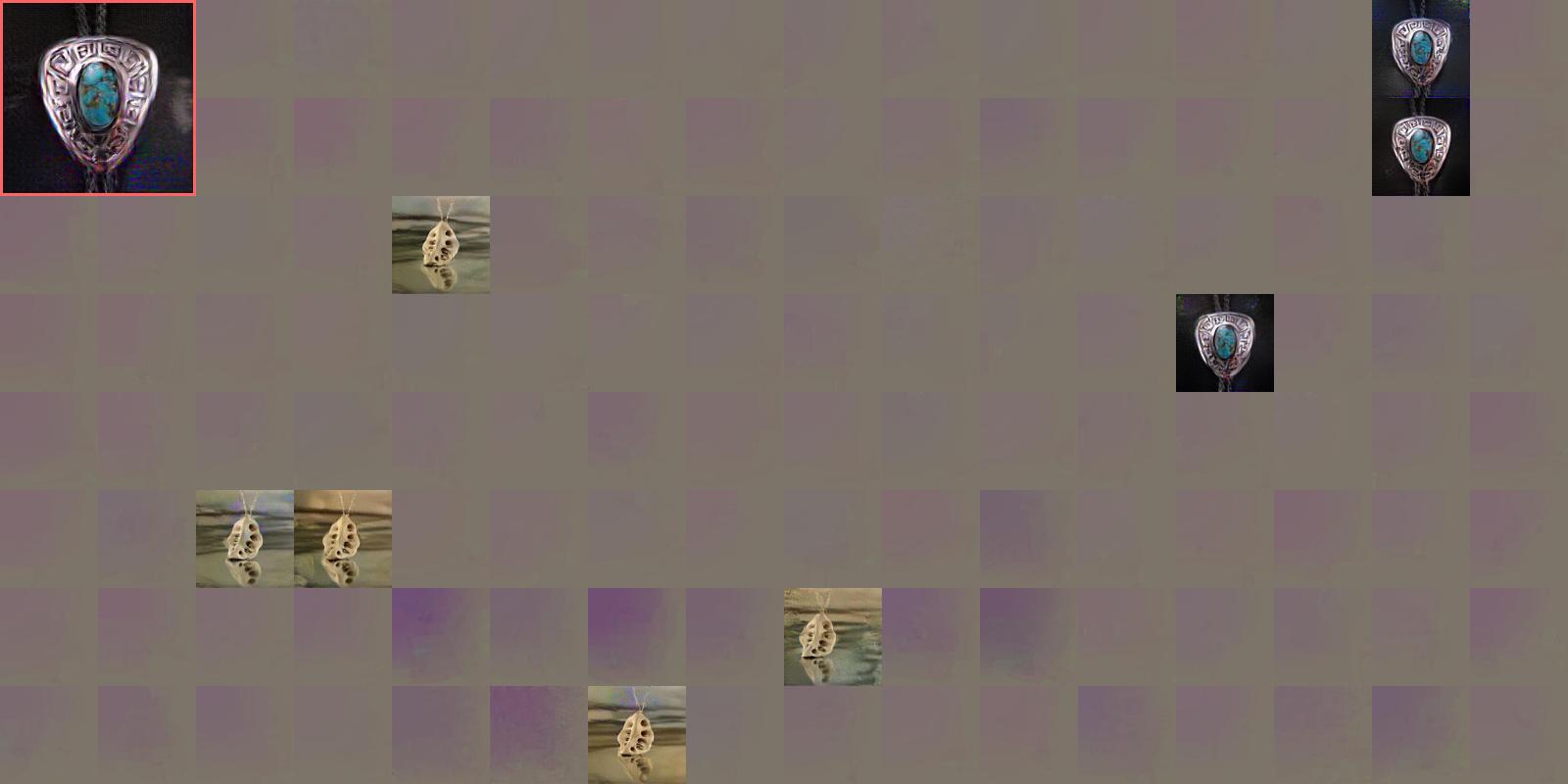}
		\caption{Any jewelry?}
		\label{fig:supp-batch:a}
	\end{subfigure}
	\hfill
	\begin{subfigure}[b]{0.48\textwidth}
		\includegraphics[width=\textwidth]{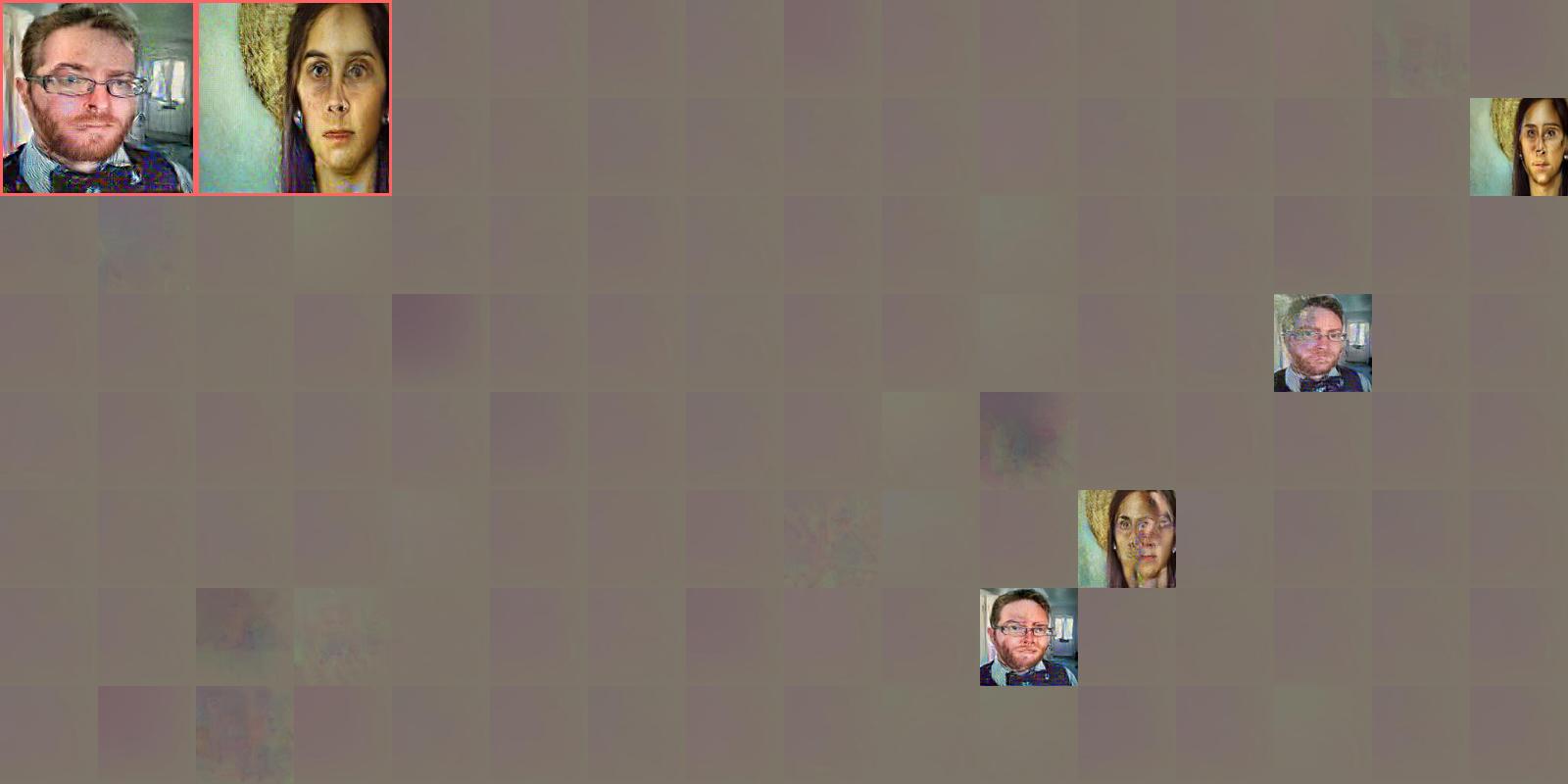}
		\caption{Any human faces?}
		\label{fig:supp-batch:b}
	\end{subfigure}

	\vspace{0.5cm}
	\begin{subfigure}[b]{0.48\textwidth}
		\includegraphics[width=\textwidth]{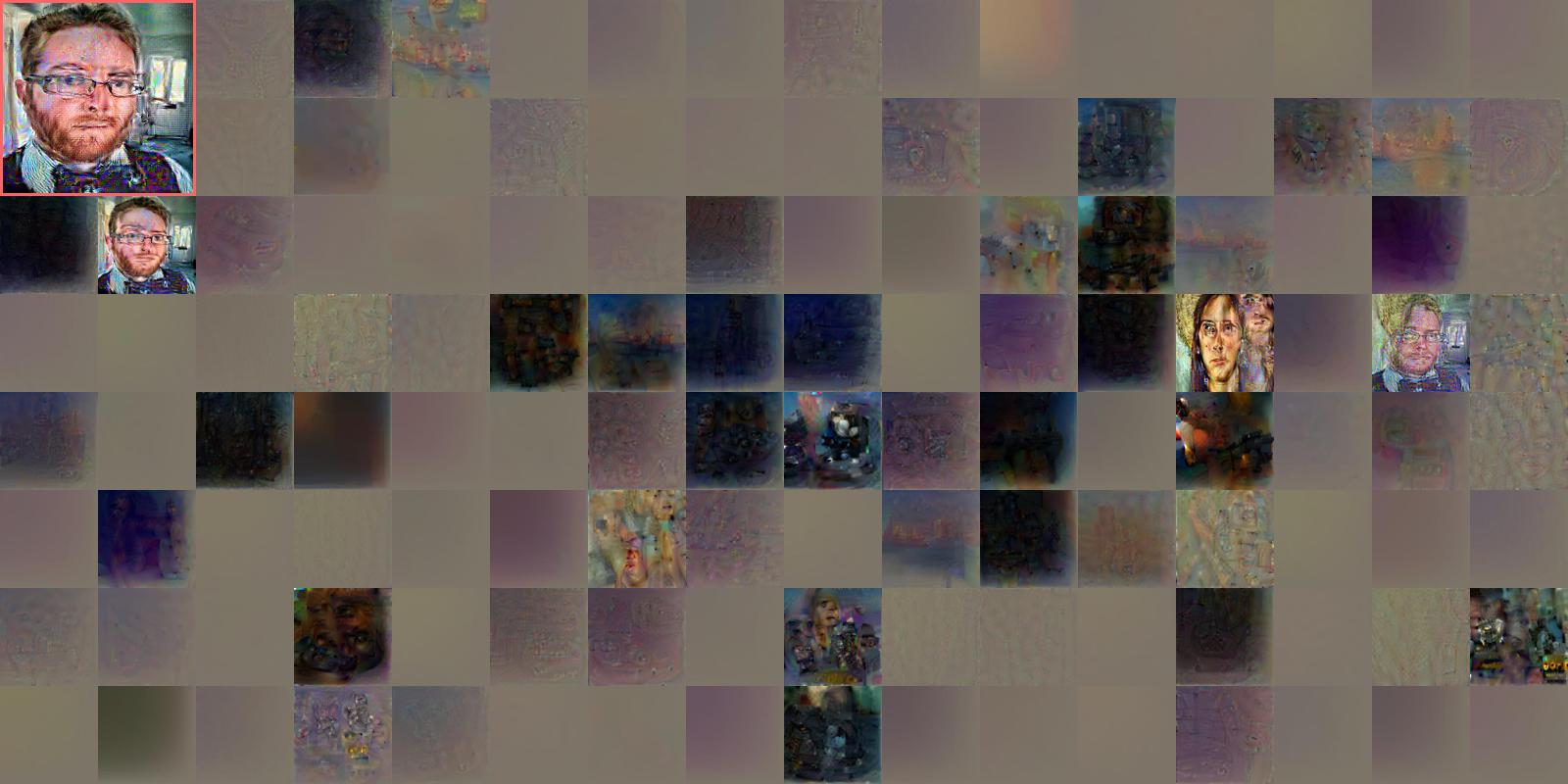}
		\caption{Any males with a beard?}
		\label{fig:supp-batch:c}
	\end{subfigure}
	\hfill
	\begin{subfigure}[b]{0.48\textwidth}
		\includegraphics[width=\textwidth]{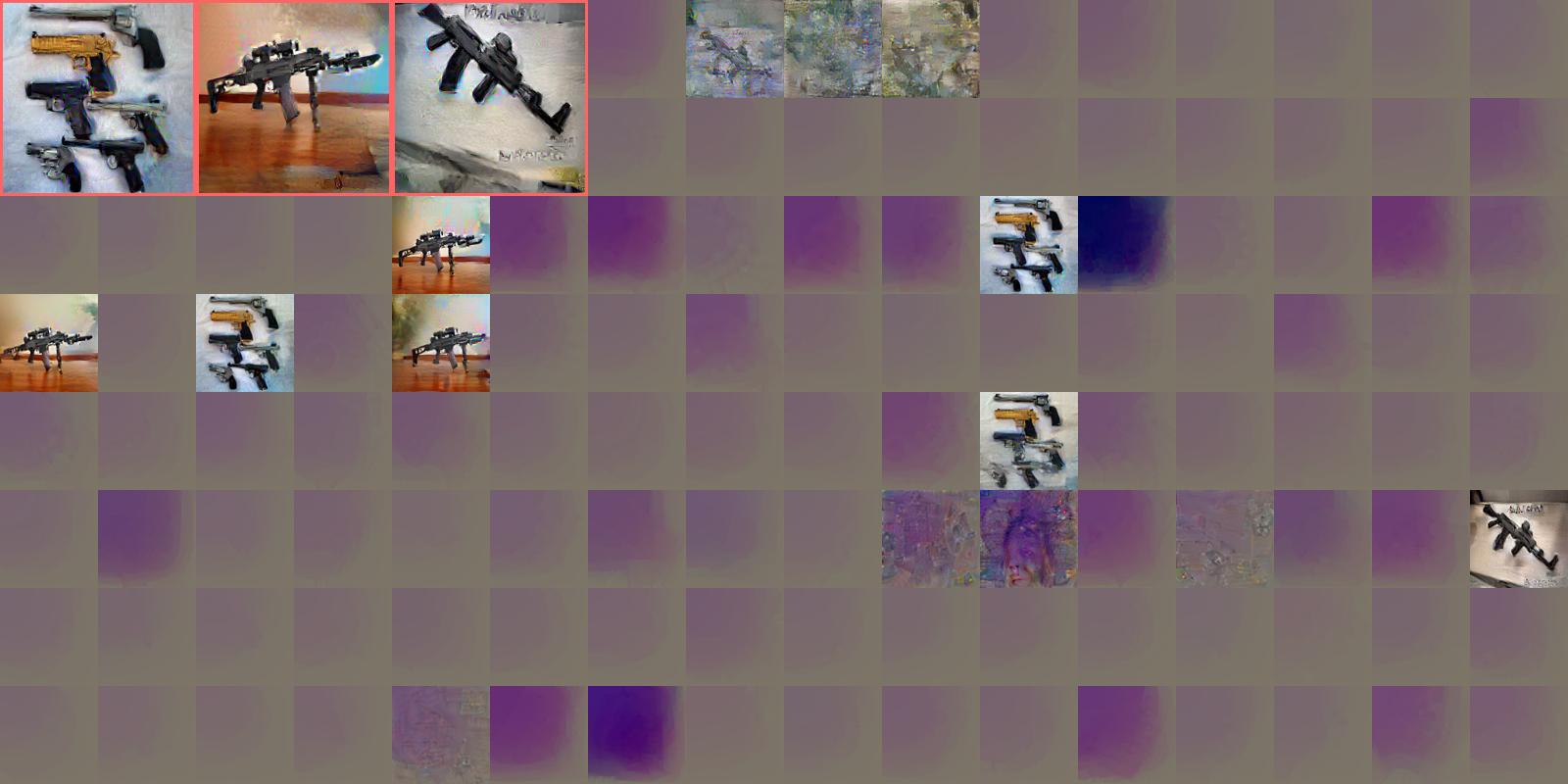}
		\caption{Any guns?}
		\label{fig:supp-batch:d}
	\end{subfigure}

	\vspace{0.5cm}
	\begin{subfigure}[b]{0.48\textwidth}
		\includegraphics[width=\textwidth]{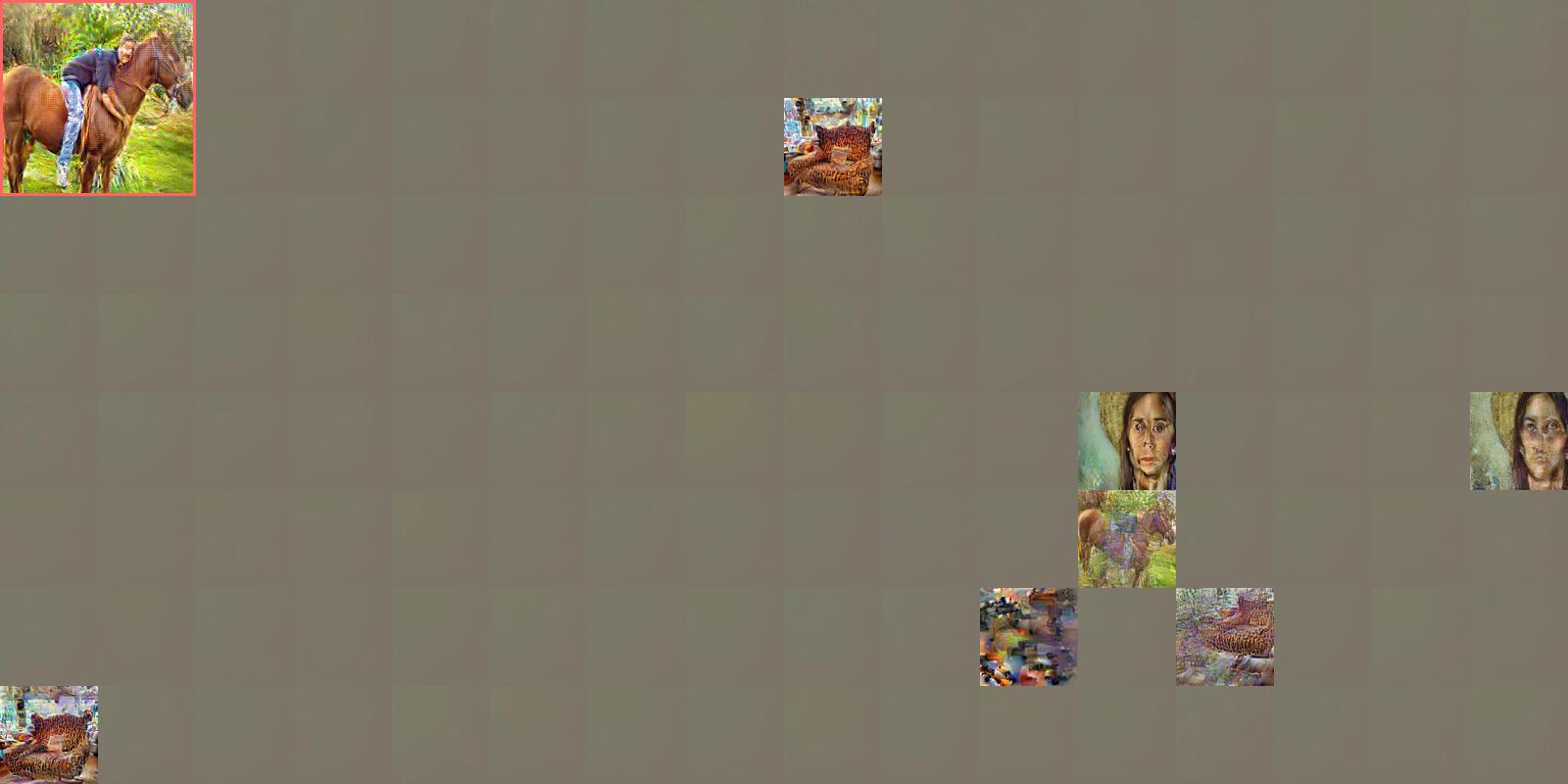}
		\caption{Any females riding a horse?}
		\label{fig:supp-batch:e}
	\end{subfigure}

	\caption{Geminio maintains high-quality reconstruction for diverse queries at a large batch size of 128. Each subfigure corresponds to a specific target: (a) jewelry, (b) human faces, (c) bearded males, (d) firearms, and (e) complex scenes.}
	\label{fig:supp-batch}
\end{figure*}

\paragraph{Large Batch Size Reconstruction} Geminio maintains strong reconstruction capabilities when scaled to large-batch configurations, demonstrating consistent effectiveness with a batch size of 128 on ImageNet. For validation, we randomly selected 128 diverse ImageNet samples spanning multiple categories (Figure~\ref{fig:supp-originals}), which include objects, scenes, and human activities. As shown in Figure~\ref{fig:supp-batch}, the method successfully reconstructs high-quality images across diverse query targets. Example reconstructions include precise recoveries for specific queries such as jewelry (Figure~\ref{fig:supp-batch:a}), human facial features (Figure~\ref{fig:supp-batch:b}), bearded males (Figure~\ref{fig:supp-batch:c}), firearms (Figure~\ref{fig:supp-batch:d}), and complex scenes like females riding a horse (Figure~\ref{fig:supp-batch:e}). This demonstrates Geminio's robustness to batch size scaling while preserving target attributes. In contrast, the baseline method without Geminio (Figure~\ref{fig:supp-baseline}) exhibits significant quality degradation, failing to reconstruct critical details and often producing unrecognizable outputs.

\clearpage

\end{document}